\documentclass[11pt]{article}

\usepackage[final]{acl}

\usepackage{times}
\usepackage{latexsym}

\usepackage[T1]{fontenc}

\usepackage[utf8]{inputenc}

\usepackage{microtype}

\usepackage{inconsolata}

\usepackage{graphicx}
\usepackage[most]{tcolorbox}
\tcbuselibrary{skins}
\newtcolorbox{promptbox}[1]{
    enhanced,
    boxrule = 1pt,
    fonttitle = \bf\color{white},
    left=2pt,
    right=2pt,
    top=2pt,
    bottom=2pt,
    colback=white,
    colframe=black,
    colbacktitle=black,
    title = #1,
    breakable
}

\usepackage{graphicx}
\usepackage{subcaption}
\usepackage{booktabs} 
\usepackage{listings}
\usepackage{xcolor}
\usepackage[T1]{fontenc}
\usepackage[utf8]{inputenc}
\lstdefinelanguage{json}{
    basicstyle=\ttfamily\footnotesize,
    breaklines=true,
    breakatwhitespace=false,
    tabsize=1,
    showstringspaces=false
}
\usepackage{placeins}
\usepackage{tcolorbox}
\tcbuselibrary{breakable}
\usepackage{caption}
\usepackage{enumitem}

\title{Towards Natural Personalization: Evaluating Long-Horizon \\ Preference Following in Personalized User-LLM Interactions}

\author{
\textbf{Qianyun Guo\textsuperscript{1}},
 \textbf{Yibo Li\textsuperscript{1}},
 \textbf{Yue Liu\textsuperscript{1}},
 \textbf{Bryan Hooi\textsuperscript{1}}
 \\
  \textsuperscript{1}National University of Singapore
  \\
  \texttt{qianyun.guo@u.nus.edu, bhooi@comp.nus.edu.sg}
}

\begin{document}
\maketitle
\begin{abstract}
Large Language Models (LLMs) are increasingly serving as personal assistants, where users may share individual preferences over extended interactions. However, assessing how well LLMs can follow these preferences in natural, long-term situations remains underexplored. This work proposes \textbf{\textsc{RealPref}}\footnote{Code and data are available at \url{https://github.com/GG14127/RealPref}.}, a benchmark for evaluating natural preference-following in personalized user-LLM interactions. \textsc{RealPref} features 100 synthetic user profiles, 1300 personalized preferences, 4 types of preference expression (from explicit to implicit), and long-horizon interaction histories. It explored three types of test tasks (multiple-choice, true-or-false, and open-ended), with granular rubrics for LLM-as-a-judge evaluation. Results indicate that LLM performance drops significantly as context length grows and preference expression becomes more implicit, and that generalizing user preference understanding to unseen scenarios poses further challenges. \textsc{RealPref} and these findings provide a foundation for future research to develop user-aware LLM assistants that better adapt to individual needs. 
\end{abstract}

\section{Introduction}
\begin{figure}[t]
    \centering
    \includegraphics[width=\columnwidth]{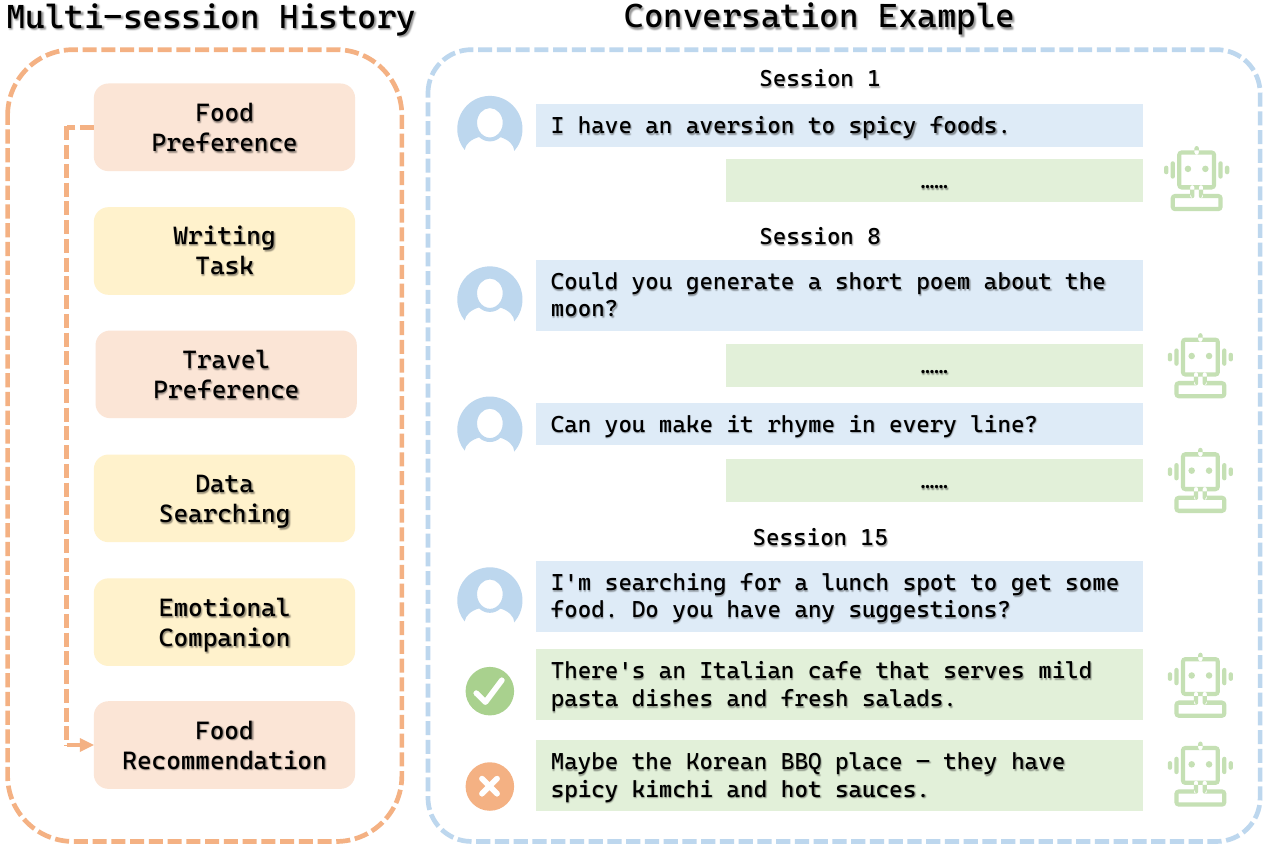}
    \caption{An example of user-LLM interaction. The conversation consists of several sessions on various topics. The user expresses a food preference in a previous session and then, after several sessions, asks the LLM for food-related recommendations.}
    \label{fig:intro}
\end{figure}

Large Language Models (LLMs) have demonstrated remarkable performance across a wide range of general-purpose tasks, such as mathematical reasoning \cite{wang2025survey}, question answering \cite{touvron2023llama, achiam2023gpt}. However, beyond serving as general-purpose assistants that support users in everyday contexts, there is a growing need for LLMs to support personalized needs, such as recommendation and decision-making. As users engage in long-term interactions with LLMs, they naturally reveal personal information, such as preferences and habits, through various forms of expression. In turn, users expect assistants to adapt to such signals and provide tailored responses to serve individual needs. Figure~\ref{fig:intro} illustrates a scenario example of user-LLM interaction that LLMs would encounter in the real world.

In light of this, we aim to answer: Can LLMs effectively (1) \textbf{capture} user preferences from complex user expressions, (2) \textbf{retain and follow} user preferences across extended interaction contexts, and (3) \textbf{generalize} user preference understanding to unseen scenarios? 

Despite rising interest in LLMs' personalization, most efforts focus on simplified scenarios:
\begin{enumerate}[itemsep=1pt,topsep=2pt,parsep=2pt]
    \item \textbf{Simple Context}. Existing works typically feature short-term context with few dialogue turns and center on a single preference or personalization instruction~\cite{salemi2023lamp,kumar2024longlampbenchmarkpersonalizedlongform,zhao2025llmsrecognizepreferencesevaluating} rather than a simulated user, resembling instruction-following instead of genuine personalization emerging from long-term interactions. 
    \item \textbf{Straightforward Preference Expressions}. While natural preference expressions could be implicit and scattered (e.g., in emotional cues), most studies assume explicit, immediately accessible traits~\cite{jiang2025knowmerespondme}, or neglect the task of generalizing preference to unseen scenarios~\cite{zhao2025llmsrecognizepreferencesevaluating}. 
    \item \textbf{Limited Evaluation Methodology}. Prevalent classification tasks diverge from natural user-LLM interactions~\cite{jiang2025knowmerespondme}. Generation tasks, though realistic, lack granular answer assessment criteria~\cite{zhao2025llmsrecognizepreferencesevaluating}.
\end{enumerate}

This work proposes \textsc{RealPref}, a benchmark that features 
diverse user biographies, varying levels of preference expressions, and comprehensive evaluation protocols, constructing a testbed that simulates the complexities of natural, long-horizon user-LLM interactions.
The main contributions of our work are summarized below:
\begin{enumerate}[itemsep=1pt,topsep=2pt,parsep=2pt]
    \item \textbf{Dataset Construction.} We constructed a user-centered long-horizon dialogue dataset that incorporates diverse preference expressions (from explicit to implicit) and generalization scenarios to simulate the complexity of natural user-LLM interactions.
    \item \textbf{Evaluation Framework.} We explored a multi-faceted evaluation protocol consisting of multiple-choice, true-or-false, and open-ended tasks, supported by granular, comprehensive rubrics to assess open-ended answers.
    \item \textbf{Empirical Findings.} Extensive experiments reveal that LLM performance declines with longer context and more implicit preferences, and that generalizing user preferences to unseen scenarios poses additional challenges.
\end{enumerate}
\section{Related Work}

\paragraph{Long-Context Memory Benchmarks.}
Early benchmarks focused on direct information retrieval, exemplified by ``Needle in a Haystack'' tests~\cite{yen2024helmet,nelson2024needle,liu2024lost}. Recent efforts have shifted towards ultra-long multi-turn interactions~\cite{tavakoli2025beyond}, such as LoCoMo~\cite{maharana2024evaluating} and Longmemeval~\cite{wu2024longmemeval}, supporting sessions exceeding 100k tokens.
Beyond simple fact recall, our work targets the active inference of user-implied preferences and their integration into ongoing interactions to provide customized responses.

\paragraph{LLM Personalization Benchmarks.}
Benchmarks in this domain typically assess (1) \textbf{Persona Adoption}: role-playing~\cite{wang2024rolellmbenchmarkingelicitingenhancing} or mimicking styles~\cite{salemi2023lamp, kumar2024longlampbenchmarkpersonalizedlongform}; (2) \textbf{User Adaptation}: customizing user-specific outputs for question-answering~\cite{zhao2025llmsrecognizepreferencesevaluating, jiang2025knowmerespondme} or downstream recommendation tasks~\cite{lyu2024llmrecpersonalizedrecommendationprompting}. Our work targets the latter for direct question-answering. Previous works like \textsc{PrefEval}~\cite{zhao2025llmsrecognizepreferencesevaluating} center dialogues on a constrained set of standalone preferences instead of long-horizon user-LLM interaction, with binary scoring criteria for open answers. \textsc{PersonaLens}~\cite{zhao2025personalens} introduces interactive, gym-like environments targeting task-oriented personalization. \textsc{PersonaMem}~\cite{jiang2025knowmerespondme} incorporates multi-session interaction histories, but overlooks implicit preferences and lacks evaluation for the free-form generation task that is close to natural interaction settings.
In contrast, \textsc{RealPref} provides a user-centered dialogue dataset featuring preference expressions ranging from explicit to implicit, supported by a granular, multidimensional rubric for evaluating open answers.
\begin{figure*}[t]
    \centering
    \includegraphics[width=0.9\linewidth]{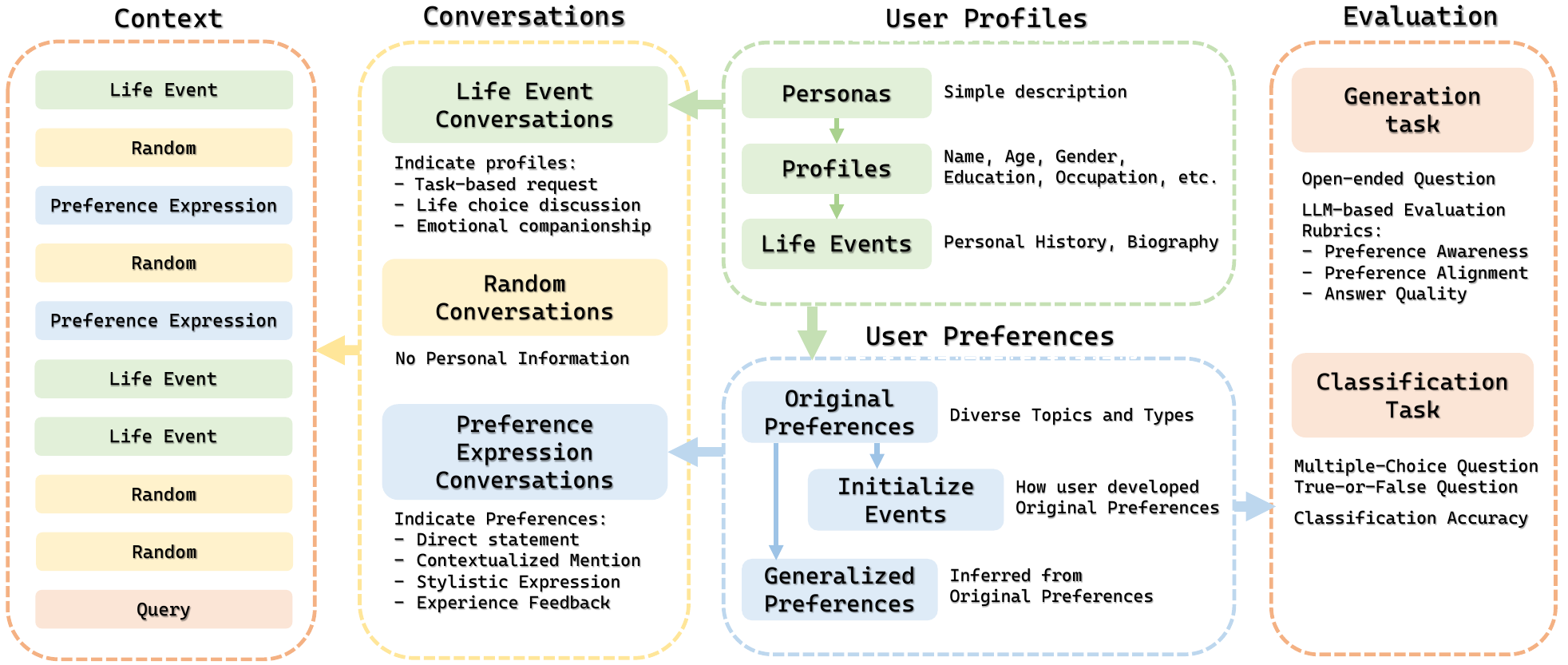}
    \caption{Generation Pipeline Overview. Starting from user personas, we construct detailed user profiles and biographies with diverse preferences. Using these, we construct rich conversation sessions that cover preference expressions from explicit to implicit, and concatenate them into a long-horizon context for each user. For each preference, generation and classification task queries are designed, with corresponding evaluation protocols.}
    \label{fig:gen_pipe}
\end{figure*}

\section{\textsc{RealPref} Dataset}

\paragraph{Problem Formulation.} To systematically evaluate preference following over extended interactions, we first establish a formal representation of the interaction data. Let $C$ denote the conversation history for a user with an LLM, which consists of several sessions $C=\{S_1, S_2, ..., S_n\}$. Each session denotes a complete dialogue segment centered on a specific topic, which may or may not indicate user information. A session consists of multiple turns: $S_i=\{(u_1^i, a_1^i), (u_2^i, a_2^i), ..., (u_m^i, a_m^i)\}$, and each turn in the session contains a pair of messages from the user and the LLM assistant. Each user has a preference set $P=\{p_1, p_2,...,p_k\}$, and the preference is indicated in a certain session $S$ in the history. For a preference $p$, a corresponding query $q$ is designed to test LLMs' understanding of that preference. At the end of the conversation, the query $q$ for target preference $p$ will be presented to the test model to obtain a response for assessment.

\paragraph{Dataset Overview.} Drawn inspiration from \textsc{PersonaMem}\cite{jiang2025knowmerespondme}, we follow the multi-stage data-generation pipeline powered by GPT-4.1 \cite{openaigpt41}, as illustrated in Figure~\ref{fig:gen_pipe}. \textsc{RealPref} consists of 100 synthetic user profiles, each with a persona, demographics, and a biography. Each user has 10 original preferences and 3 generalized preferences that can be rationally inferred from the original ones, constructing 1300 preference-query pairs in total. The query has three forms: multiple-choice, true-or-false, and open-ended. Regarding conversations, there are 4 types of preference expressions for each original preference, and 10-15 life-event conversation sessions per user. In total, each user has a highly personalized context of up to 2-3K tokens. Additionally, we have around 1,000 turns of user-irrelevant random conversations for context length control. The following sections will use Figure~\ref{fig:gen_pipe} as a roadmap to introduce our work in the following order: Profiles, Preferences, Conversations, Context, and Evaluation. Examples and prompts for each dataset part are provided in Appendix~\ref{app:dataset}, and detailed benchmark-quality verification steps are provided in Appendix~\ref{app:quality}.

\subsection{Profiles} 
Previous work in RecSys~\cite{pazzani1999framework,adomavicius2010context} has shown that a user's demographic characteristics and relevant contextual information, such as current life situations, may affect their preferences. Our design aims to generate diverse, user-specific preferences based on varied user backgrounds and life situations. We first sampled and manually filtered 100 random \textit{\textbf{Personas}} from PersonaHub \cite{ge2025scalingsyntheticdatacreation}, a collection of 1 billion diverse personas (brief sentences summarizing personal traits) from web data. Persona samples are then augmented with additional demographic details and descriptions to form complete \textit{\textbf{Profiles}}.
Furthermore, we constructed a \textit{\textbf{Life Events}} history for each user, including life milestones like graduation. These diverse user backgrounds and life situations will help foster varied preferences and natural, personalized, long-term conversations.

\subsection{Preferences}
\paragraph{Original Preferences.} ``Original Preferences" here refers to preferences generated directly from user profiles.
Each original preference involves an \textit{\textbf{Initialize Event}} that records when and how the user developed the preference.

High-quality preference data is essential for a robust evaluation. If preferences are overly generic or predictable—like ``disliking fast food"—the task becomes trivial for LLMs: They may pass the test by relying on common sense instead of user-specific context. Therefore, we highlight requirements to ensure the generated preferences are \textit{Diverse}, \textit{Persona-Related}, \textit{Unique}, \textit{Value-Free}, and \textit{Complete}, with details in Appendix \ref{app:pref}.
Additionally, we introduced a \textit{Topic Pool} and defined our \textit{Preference Types} to ensure preference diversity:
\begin{itemize}[itemsep=1pt,topsep=2pt,parsep=2pt]
    \item \textbf{Preference Topics.} We created a topic pool comprising 7 broad categories and 37 fine-grained subtopics. (See Appendix \ref{app:pref} Table \ref{tab:topic_pool}.) For each user, we both randomly sample topics and let the LLM select additional ones, spreading topics widely while tailoring generation to individual traits. 
    \item \textbf{Preference Types.} To capture the nuances of human preferences, we categorize them into types listed in Table \ref{tab:pref_type} based on their descriptive nature, which will help establish criteria for evaluating open-ended responses.
\end{itemize}

\begin{table}[t]
\centering
\begin{tabular}{cl}
\toprule
\multicolumn{1}{c}{\textbf{Type}}   & \multicolumn{1}{c}{\textbf{Description}}                                                                       \\ \midrule
\begin{tabular}[c]{@{}c@{}}\textbf{Relative} \\ \textbf{Preference} \end{tabular}  & Prefer something over another                                                                                  \\ \hline
\begin{tabular}[c]{@{}c@{}}\textbf{Negative} \\ \textbf{Preference} \end{tabular}  & \begin{tabular}[c]{@{}l@{}}Dislike or aversion to something \\ (Anything else is acceptable)\end{tabular} \\ \hline
\begin{tabular}[c]{@{}c@{}}\textbf{Exclusive} \\ \textbf{Preference} \end{tabular}  & \begin{tabular}[c]{@{}l@{}}Only like or accept something\\  (Anything else is rejected)\end{tabular}     \\ \hline
\begin{tabular}[c]{@{}c@{}}\textbf{Positive} \\ \textbf{Preference} \end{tabular}  & \begin{tabular}[c]{@{}l@{}}Like or enjoy something\\ (Without aversion to alternatives)\end{tabular}\\ \bottomrule
\end{tabular}
\caption{Preference Types Classified by Descriptions.}
\label{tab:pref_type}
\end{table}

\paragraph{Generalized Preferences.} The term ``Generalized Preferences" refers to preferences that differ from the original preferences but can be rationally deduced from them.
Naturally, people may have linked preferences across domains, driven by similar reasons~\cite{fernandez2012cross}. For example, someone who prefers cruelty-free, eco-friendly beauty products may also favor natural, eco-friendly cleaning supplies, reflecting a holistic lifestyle. When users state a preference with reasons, they may expect the assistant to recognize their general tendency across topics, similar to how recommendation systems suggest new products based on user history. Therefore, only original preferences will be tested with expression conversations in the context, while generalized preferences will be tested directly.

\subsection{Conversations} 
Conversation sessions are generated from curated user profiles and preferences to simulate daily dialogue data. In real-world user-LLM interactions, users do not always discuss preferences~\cite{zhao2024wildchat}; they may instead discuss their lives or be simply task-oriented, revealing no personal information. Depending on the type of user information indicated, 3 categories of conversations are generated to support long-term interactions.

\paragraph{Life-Event Conversation.} This category indicates user profiles and life events. Users' life event histories comprise distinct life stages, each of which requires different assistance from LLMs. Typical content includes task-based requests, life-choice discussions, and emotional companionship, through which the user gradually reveals their background or circumstances.
 
\paragraph{Random Conversation.} This category will not reveal any user-specific information. Typically, the user begins with a specific request, such as factual questions, troubleshooting, or small talk. Given their user-free nature, random conversations are suitable for sharing among users and for managing context length in experiments.

\paragraph{Preference-Expression Conversation.} This category indicates user preferences and plays a crucial role in our work. Naturally, in addition to explicit statements, users may express their preferences implicitly through feedback or emotional cues. Therefore, we provide four \textit{Preference Expression Types} from explicit to implicit, which require varying levels of capability, ranging from basic information extraction to complex comprehension and reasoning. Descriptions of each type are provided below:
\begin{enumerate}[itemsep=1pt,topsep=2pt,parsep=2pt]
    \item \textbf{Explicit - Direct Statement.} Preferences are stated directly and explicitly in a single turn without redundant words.
    
    \item \textbf{Explicit - Contextualized Mention.} In multi-round conversations, users start the session with other related daily topics as context instead of directly stating preferences. As the dialogue advances, preferences are naturally mentioned explicitly in the context.
    
    \item \textbf{Implicit - Stylistic Expression.} The context resembles the second type, but users will not state preferences with explicit words like Like, Prefer, or Dislike. Instead, they imply preferences through stylistic devices such as rhetorical questions, metaphors, contrasts, or emotional expression. 
    
    \item \textbf{Implicit - Experience Feedback.} This type involves multiple sessions over time, based on the user's \textit{Preference Initialize Event}. The user won't explicitly state their preference in one session but will reveal it gradually through experience sharing or feedback across sessions. Preferences can only be inferred by integrating information from all sessions.

    For example, the early session may occur before the event, when users are willing to try without preference. After the event, users may share their experiences, give feedback, and indicate preferences.
\end{enumerate}

\subsection{Contexts}
Based on the generated conversation sessions, we can build a complete interaction history context for each user. The context includes all user-specific sessions (life-event and preference-expression conversations) in chronological order, with random conversations inserted. Two insertion methods are used: one randomly inserts into existing history to dilute personal information, and the other appends it at the end, distancing preference cues from the query. Table \ref{tab:context} lists the context configurations we provided for the following experiments. 

\begin{table}[t]
\centering
\begin{tabular}{lccc}
\toprule
\textbf{Context}            & \textbf{Interval} & \textbf{Tail} & {\textbf{Total Length}} \\ \midrule
\textbf{Simple}            & -- & -- & 2K\\
\textbf{Normal}             & 35K & -- & 37K\\
\textbf{Long}               & 70K & -- & 72K\\
\textbf{Very Long}          & 140K & -- & 142K\\
\textbf{Extended}    & 35K & 105K & 142K\\
\textbf{Extreme} & 35K & 210K & 247K\\ \bottomrule
\end{tabular}
\caption{Context Configurations. \textit{Interval} Tokens set the length of interval-inserted random sessions. \textit{Tail} Tokens set the length of random sessions added to the end.}
\label{tab:context}
\end{table}
\section{\textsc{RealPref} Evaluation}

\subsection{Task Types} 
Based on the generated preferences, we provide classification (multiple-choice, true-or-false) and generation (open-ended) tasks. 
See query examples in Appendix~\ref{app:query} and verification steps in Appendix~\ref{app:infer}.
 
\textbf{Open-Ended Questions.} The generation task is the closest to the real-world natural user-LLM interaction setting. High-quality queries must be discriminative: answering without knowledge of preferences may violate them~\cite{zhao2025llmsrecognizepreferencesevaluating}. Thus, query content should not reveal user preferences, preventing LLMs from relying solely on common-sense reasoning to respond accurately.

\textbf{Multiple-Choice Questions (MCQs).} MCQ is the most prevalent classification task in previous works~\cite{zhao2025llmsrecognizepreferencesevaluating, jiang2025knowmerespondme}. The queries here in the MCQs are identical to the open-ended ones. Four options are generated, all relevant and reasonable answers to the query, but only one best aligns with the preference.

\textbf{True-or-False Questions.} We suspect that LLMs can answer MCQs correctly by comparing the options, even without knowledge of the user's preference, so we designed a mirror classification task. A true-or-false question combines the identical query with one option from the MCQ, asking whether the option is suitable. In this setting, LLMs decide based on a single option without option comparison. We construct a correct and an incorrect version of each query, yielding a balanced dataset.

\subsection{Evaluation Methods} 
While classification tasks adopt accuracy as the metric, the generation task requires a comprehensive set of evaluation criteria to assess response quality. Based on prior work \cite{zhao2025llmsrecognizepreferencesevaluating} and an analysis of actual responses, this work provides a more granular scoring criterion on a 1-5 scale from 3 rating dimensions: \textit{Preference Awareness}, \textit{Preference Alignment}, and \textit{Answer Quality}. See detailed evaluation rubrics in Appendix~\ref{app:rubrics} and verification steps in Appendix~\ref{app:veriscore}.

\textbf{Preference Awareness.} The first dimension focuses on the extent to which the answers recognize or understand user preferences. The judgment is based on the following: If an LLM is aware of user preferences when responding, it would mention the preferences it uses in the answer.

\textbf{Preference Alignment.} The second dimension focuses on how well the answer's content aligns with user preferences by checking if the answer's actual content aligns with user preferences, with criteria discussed by preference types in Table~\ref{tab:pref_type}.

\textbf{Answer Quality.} The last dimension focuses on the extent to which the answer's content is actually helpful and constructive to the user. Although not directly tied to preferences, it is a crucial factor for assessing the quality of open-ended responses. 
\section{Experiments}
\subsection{Experiment Settings}
\begin{figure}[ht]
    \centering
    \includegraphics[width=\columnwidth]{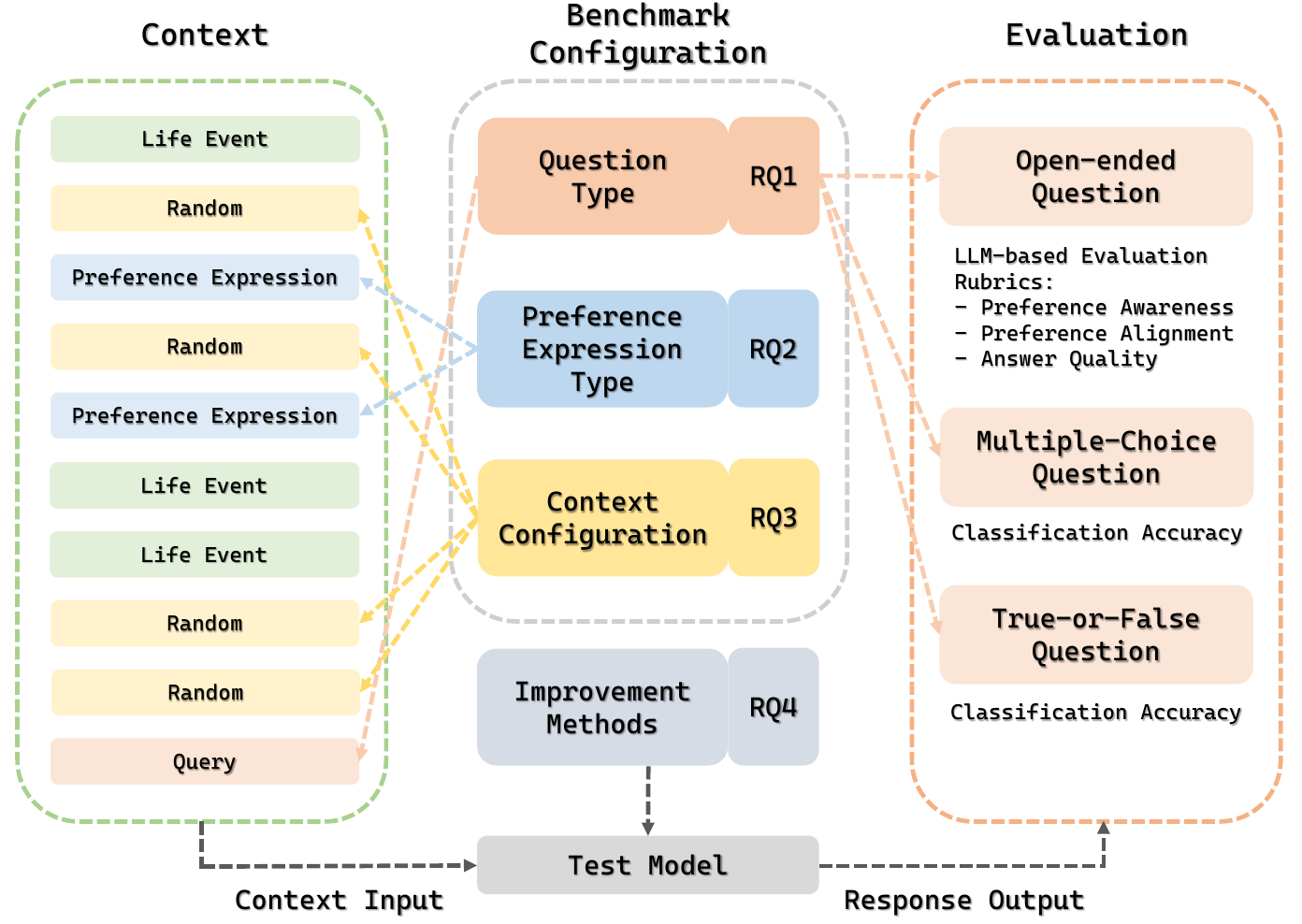}
    \caption{Benchmark Configuration Overview. Question Type (\textit{Multiple-Choice}, \textit{True-or-False}, and \textit{Open-Ended}) sets the test and evaluation; Preference Expression Type (\textit{Direct Statement}, \textit{Contextualized Mention}, \textit{Stylistic Expression}, and \textit{Experience Feedback}); Context Configuration sets the history input; Improvement Methods (\textit{Reminder}, \textit{Few-Shot CoT}, and \textit{RAG}) are tested with models. Each factor above is related to a research question (RQ) we will explore in the experiments.}
    \label{fig:exp}
\end{figure}

\paragraph{Benchmark Configurations.} Our experiments explore the performance of current LLMs across several dimensions. Figure~\ref{fig:exp} illustrates an overview of different benchmark configurations and their related control components in the experiment: \textit{Question Type} sets the test task; \textit{Preference Expression Type} controls the preference expression sessions included in the context; \textit{Context Configuration} controls the context length by managing the insertion of random sessions.

\paragraph{Base Models.} We chose the following LLM base models for experiment: \textit{GPT-5}, \textit{GPT-5 mini}, \textit{Qwen3-235B-A22B Instruct}, \textit{Gemini 2.5 Flash-Lite}, \textit{Llama 3.3 70B Instruct}, \textit{DeepSeek-V4-Pro}, \textit{Qwen3-8B}. For open-ended answers, we specify \textit{Claude Sonnet 4} as our LLM judge. See Appendix~\ref{app:models} for model details.

\paragraph{Methods.} We test models zero-shot by default, but we also evaluate several basic improvement methods: (1) \textbf{Reminder}: Add a sentence reminding the model to review the user's past preferences in response. (2) \textbf{Few-shot Chain-of-Thought}: For open-ended questions, give 3 examples with excellent responses that mention user preferences, include detailed chain-of-thought, and offer specific guidance. (3) \textbf{Retrieval-Augmented Generation (RAG)}: Offer the top 5 retrieved history turns that have the highest similarity to the current question using a sentence embedding model. See Appendix~\ref{app:method} for detailed implementation.

\subsection{Research Questions}
Based on the setting factors described above, we proposed the following research questions (RQs):

\textbf{RQ1}: Which \textbf{evaluation task} is better at helping us understand the preference-following abilities of LLMs?

\textbf{RQ2}: How does LLM’s preference-following performance changes across different \textbf{preference expression} types?

\textbf{RQ3}: How does LLM’s preference-following ability perform in \textbf{long contexts}?

\textbf{RQ4}: How do different \textbf{improvement methods} affect the LLMs' ability to follow user preferences?

\textbf{RQ5}: Can LLMs \textbf{generalize} user preferences to unseen scenarios?

The subsequent experiments will focus on these questions and delve deeper into the model's performance under various conditions.

\subsection{Experiment Results and Analysis}
\subsubsection{Question Type Comparison} \label{sec:rq1}
\begin{figure}[t]
    \centering
    \includegraphics[width=0.95\columnwidth]{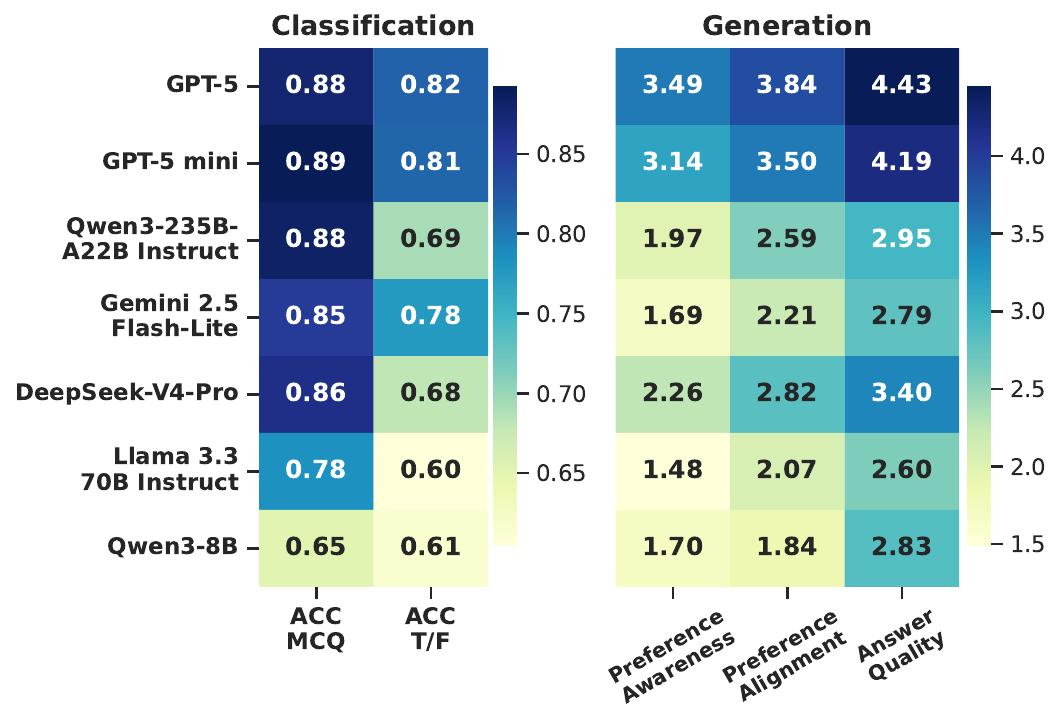}
    \caption{Model Performance across Question Types with \textit{Normal} context, \textit{Experience Feedback} expression, \textit{Zero-shot}.}
    \label{fig:rq1}
\end{figure}

To investigate \textbf{RQ1}, we test candidate LLMs in \textit{Normal} context, \textit{Experience Feedback} expression type, \textit{Zero-shot}, across three types of evaluation tasks: \textit{MCQ}, \textit{True-or-False}, and \textit{Open-ended Questions}. Figure~\ref{fig:rq1} shows the experiment results.

\textbf{For classification tasks, True-or-False questions are better than MCQs in reflecting LLMs' ability to follow user preferences.} 
As shown in Figure~\ref{fig:rq1}, MCQ scores are generally much higher than the random line at 0.25, with slight differentiation across LLMs. Despite identical queries and options from MCQs, True-or-False question scores are typically lower (closer to the random line at 0.5). The overly high scores on MCQs raise suspicions that the LLM may exploit which options appear different (i.e., 'the odd one out') to infer likely answers, even without knowledge of user preferences. We explored "Blind Test" to further address this situation (see Appendix~\ref{app:infer}). Such a situation is systematically unavoidable when the user preference belongs to \textit{Negative Preference} (Dislike or aversion to something): Since the three incorrect options share the same characteristics, the correct option is more likely to stand out (see example in Appendix~\ref{app:query}). In contrast, True-or-False questions prevent option comparison, making results more accurately reflect preference-following ability. 

\textbf{Compared to classification tasks, generation tasks are better at distinguishing between LLMs that can proactively follow user preferences and those that cannot.}
As shown in Figure~\ref{fig:rq1}, while the rankings are similar to classification tasks, the performance gap between the GPT-5 series and other models is much more pronounced in the generation task. The generation task requires proactively generating user-aligned answers that closely resemble dialogue in daily interactions, whereas the classification task is a passive selection from existing options, allowing models to match options to historical turns. We aim to establish an evaluation environment that more closely reflects everyday scenarios, and we expect the generation task to present a higher ceiling of difficulty in future work.

\begin{figure}[t]
    \centering
    \includegraphics[width=\columnwidth]{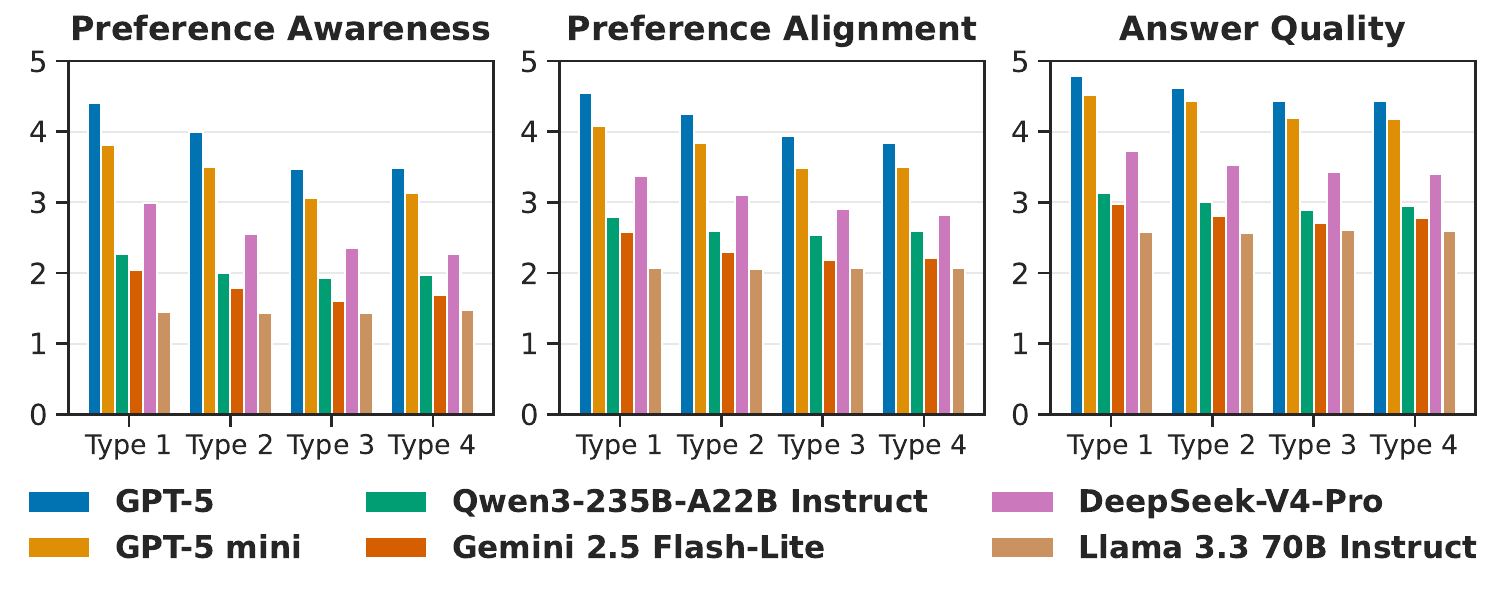}
        \caption{Model Performance across Expression Types with \textit{Normal} context, \textit{Zero-shot}. Type 1: Explicit - Direct Statement, Type 2: Explicit - Contextualized Mention, Type 3: Implicit - Stylistic Expression, Type 4: Implicit - Experience Feedback.}
        \label{fig:rq2}
\end{figure}
\begin{figure}[t]
    \centering
    \includegraphics[width=\columnwidth]{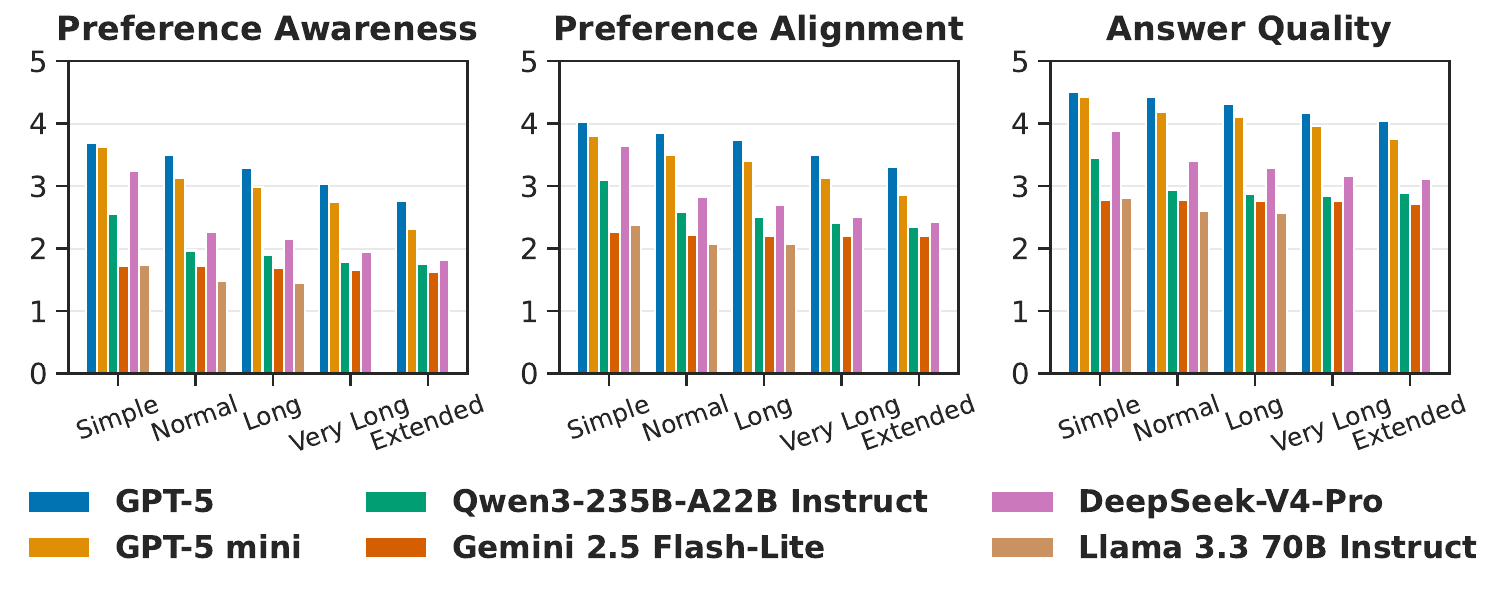}
        \caption{Model Performance across Context Configurations with \textit{Experience Feedback} expression, \textit{Zero-shot}. Context Length in tokens: Simple (2K), Normal (37K), Long (72K), Very Long (142K), Extended (142K).}
        \label{fig:rq3}
\end{figure}

\begin{figure*}[t]
 \centering
    \includegraphics[width=0.68\linewidth]{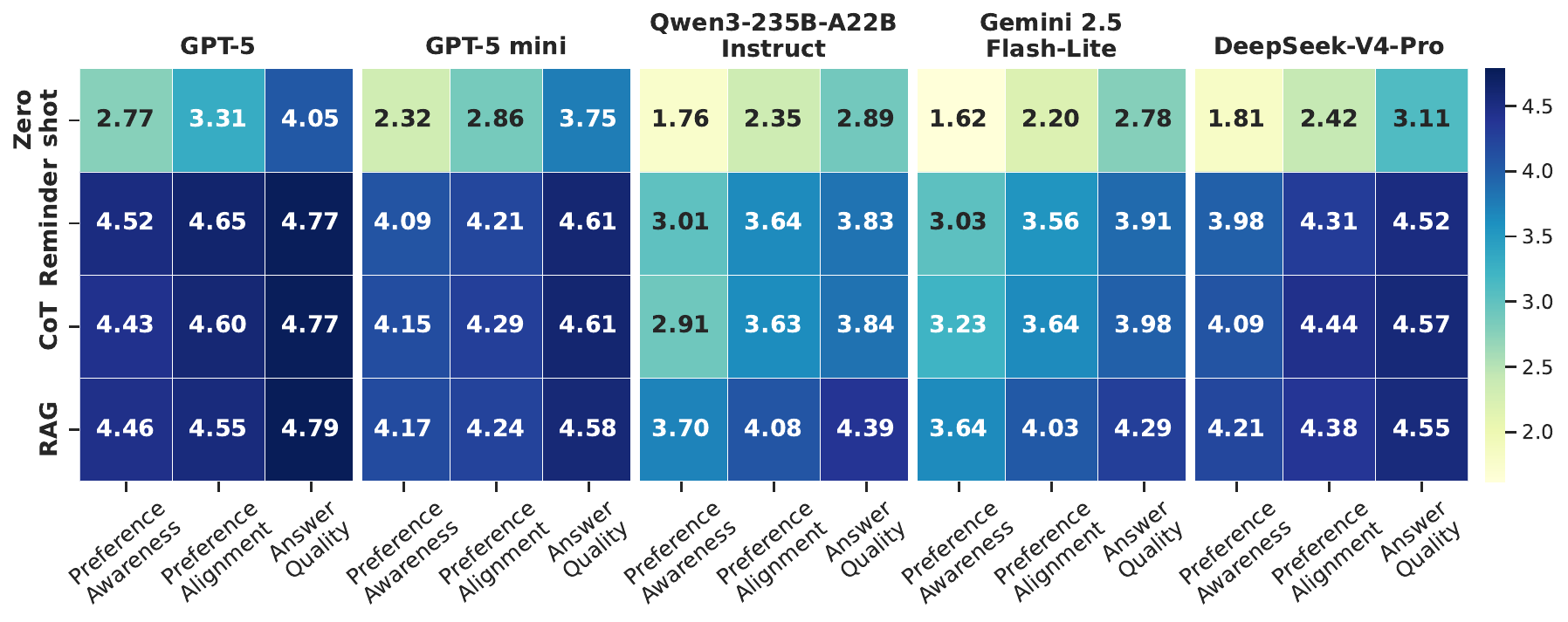} \hfill
    \includegraphics[width=0.29\linewidth]{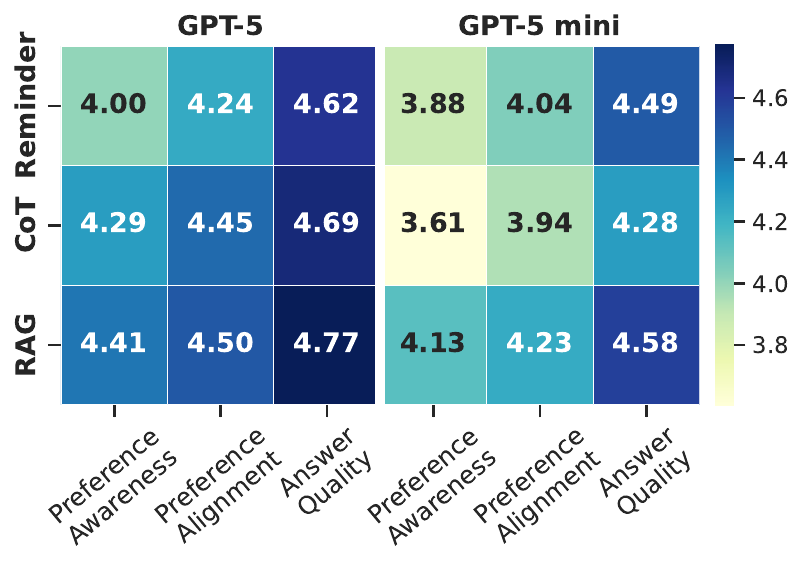}
    \caption{Model Performance across Methods with \textit{Experience Feedback} expression, under \textit{Extended (142K)} context (left) and \textit{Extreme (247K)} context (right).}\label{fig:rq4}
\end{figure*}

\subsubsection{Effect of Preference Expressions} \label{sec:rq2}
Results toward \textbf{RQ1} demonstrated the effectiveness of generation tasks. For \textbf{RQ2}, we test LLMs in \textit{Open-ended Question}, \textit{Normal} context, \textit{Zero-shot}, across four types of preference expressions. Results are shown in Figure~\ref{fig:rq2}. 

\textbf{LLMs' overall preference following ability noticeably declines from explicit expressions to implicit expressions.} Figure~\ref{fig:rq2} illustrates that the scores across all rating dimensions noticeably decline from explicit expression (Type 1 and 2) to implicit expression (Type 3 and 4), with Type 1 scoring higher than Type 2. The findings above align with expectations: \textit{Direct Statement} clearly presents the preference, making it the easiest to identify. \textit{Contextualized Mention} mixes preferences with context, posing challenges in extraction. \textit{Stylistic Expression} and \textit{Experience Feedback} further complicate the task, requiring LLMs to capture preference information beneath the expressions. This downward trend is more pronounced in GPT-5 and GPT-5 mini than in the other LLMs (significant tests reported in Appendix~\ref{app:sig}). This may be because other models do not actively retrieve or follow user preferences during free-form generation, so variations in expression have minimal impact on their performance.

\subsubsection{Preference Following in Long Contexts}

For \textbf{RQ3}, we test candidate LLMs in \textit{Open-ended Question}, \textit{Experience Feedback} expression type, \textit{Zero-shot}, across five context configuration types defined in Table~\ref{tab:context}: \textit{Simple} (2K), \textit{Normal} (37K), \textit{Long} (72K), \textit{Very Long} (142K), and \textit{Very Long Extra} (142K). Results are shown in Figure~\ref{fig:rq3}.

\textbf{LLMs' preference-following performance drops obviously as context length increases.} Figure~\ref{fig:rq3} shows a clear downward trend in rating scores as context length increases from \textit{Simple} to \textit{Very Long}, particularly in the \textit{Preference Awareness} and \textit{Preference Alignment} dimensions. Although under the same context lengths, there is a further drop from \textit{Very Long} to \textit{Extended} by changing the insertion location of random sessions from the history interval to the tail, showing that besides overall context length, the interval between useful preference cues and the final query also affects performance. Similar to the observation in RQ2, such a downward trend is more obvious in the GPT-5 series than in the other LLMs.

\subsubsection{Effect of Basic Improvement Methods}

For \textbf{RQ4}, we compared the performance of candidate LLMs in \textit{Open-ended Question}, \textit{Experience Feedback} expression type, with \textit{Extended} (142K), and \textit{Extreme} (247K) context across baseline and 3 methods: \textit{Zero-shot}, \textit{Reminder}, \textit{Few-shot Chain-of-Thought}, and \textit{Retrieval-Augmented Generation}. The experiment results are shown in Figure~\ref{fig:rq4}.

\textbf{For context length within the LLM's retrieval capability, \textit{Reminder}, \textit{Few-shot CoT} and \textit{RAG} show similar improvement effects.} In Figure~\ref{fig:rq4} (left), for the GPT-5 series, all three methods show similar improvements over the \textit{Zero-shot} setting, but \textit{Reminder} remains the most cost-effective method, delivering comparable performance.

\textbf{For longer contexts, \textit{RAG} is most effective in improving model performance.} In Figure~\ref{fig:rq4} (left), for Qwen3-235B-A22B Instruct and Gemini 2.5 Flash-Lite in the \textit{Extended} (142K) context, \textit{RAG} outperforms the other methods. Similar situations occurred for the GPT-5 series in the \textit{Extreme} (247K) context shown in Figure~\ref{fig:rq4} (right). These findings indicate the effectiveness of \textit{RAG} in enabling LLMs to achieve preference following in contexts that may exceed their retrieval capabilities.

\subsubsection{Preference Generalization}
\begin{figure}[t]
    \centering
    \includegraphics[width=\columnwidth]{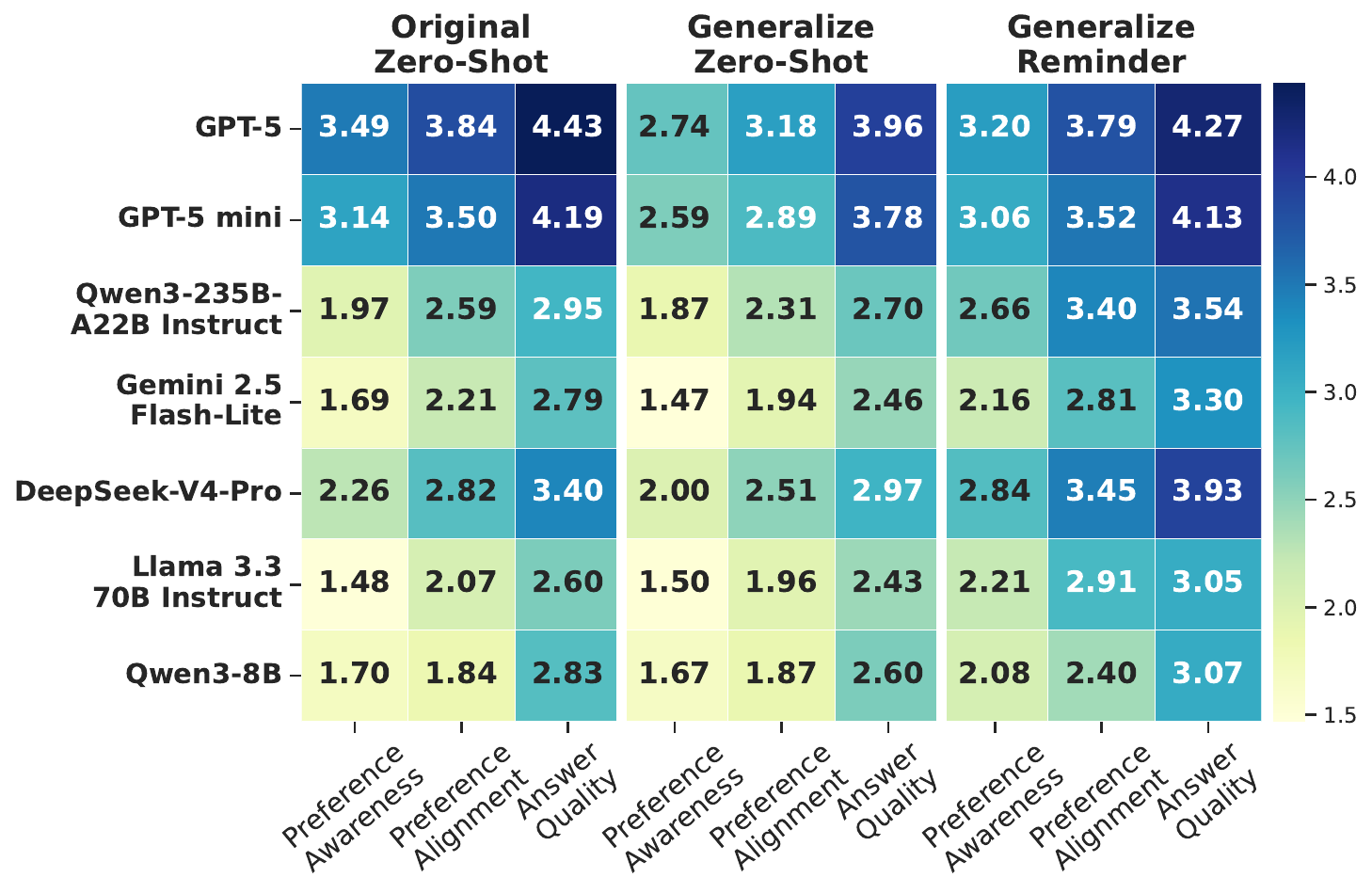}
    \caption{Model Performance for Preference Generalization with \textit{Normal} context, \textit{Experience Feedback} expression.}
    \label{fig:rq5}
\end{figure}

To explore \textbf{RQ5}, we test candidate LLMs in answering generalized preference queries in \textit{Open-ended Question}, \textit{Experience Feedback} expression type, \textit{Normal} context, with \textit{Zero-shot} and \textit{Reminder} methods. Generalized preferences differ from the original ones but can be inferred from them because they share similar underlying reasons. The contexts contain only the original preference expressions, requiring LLMs to generalize their understanding to unseen scenarios, which aligns with real-world situations and needs~\cite{fernandez2012cross}. The experiment results are shown in Figure~\ref{fig:rq5}. 

\textbf{Current LLMs lack proactivity to generalize the understanding of user preferences to new scenarios.} Figure~\ref{fig:rq5} shows that LLMs performed worse on generalized preference queries than on original ones in the same \textit{Zero-shot} setting. It is notable that for the Qwen3 series, Gemini 2.5 Flash-Lite, Llama 3.3 70B Instruct, and DeepSeek-V4-Pro, the scores for generalized preferences with \textit{Reminder} are higher than the scores for original ones under \textit{Zero-shot} setting, showing that \textit{Reminder} effectively helps stimulate these LLMs' proactivity to recall original preferences, infer generalized preferences, and give preference-aligned answers. However, for the GPT-5 series, although \textit{Reminder} improves performance, the scores for generalized preferences remain lower than those of the original ones under the \textit{Zero-shot} setting, indicating that following generalized preferences still poses more challenges for LLMs.
\section{Conclusion}
In this work, we proposed \textsc{RealPref}, a comprehensive benchmark for evaluating LLMs’ ability to follow preferences in natural, long-horizon, personalized user-LLM interactions. \textsc{RealPref} features a long-horizon conversation dataset constructed by a user-centric generation pipeline. Our design considers diverse preference topics and types, complex preference expressions, preference generalization scenarios, and explores different evaluation tasks. 
Experimental results show that many LLMs lack proactivity in recalling and following user preferences in the open-ended generation task, the task that most closely aligns with natural user-LLM interaction. Even capable models exhibit significant performance degradation as context length increases and preference expressions become more implicit. Besides, generalizing learned preferences to unseen scenarios poses additional challenges. 
\textsc{RealPref} and these findings may provide a foundation for future research to develop smarter, user-aware LLM assistants tailored to individual needs.
\section*{Limitations}
This work introduces \textsc{RealPref}, a benchmark for evaluating natural preference-following in long-horizon user–LLM interactions simulating real-world scenarios. However, it also has limitations to acknowledge. 

\textbf{Synthetic dataset.} All user profiles, preferences, and dialogue histories in our work are synthetically generated. Although this approach enables controlled evaluation and protects privacy, synthetic users may not fully capture the complexity and unpredictability of human behavior. Besides, the topic pool, preference space, and preference expressions, while diverse, remain finite and may not cover all domains, languages, or cultural contexts encountered in real-world applications. Nevertheless, we make deliberate efforts through diverse topic design and varied preference expressions to make the generated data closely approximate natural interactions. More precisely, \textsc{RealPref} should be viewed as a controlled diagnostic benchmark for long-horizon preference following, rather than a complete simulation of real user behavior.

\textbf{LLM judges in Evaluation.} Our work relies on LLM-as-a-judge with predefined rubrics. While automatic judging may not perfectly align with human judgments, we carefully design detailed, structured evaluation criteria and conduct verifications to ensure consistency and reliability. Moreover, compared with large-scale human evaluation, this approach offers a significantly more cost-efficient and scalable alternative, making large-scale systematic evaluation feasible.

\textbf{Experiments.} As a benchmark effort, the experiment part of our work is to reveal the performance of current LLMs, with the settings focused on varied benchmark configurations. Since we provided many variable settings in our benchmark, we varied the experimental variable while holding the others at representative settings to simplify the experimental procedure. Also, given the long context and API costs involved in the test, we assume that the LLM candidates and the judge used in the experiment are stable versions that yield stable test results, so the presented test results are from a single run. 

\section*{Ethical Considerations}
This work serves as a benchmark that enables a more natural assessment of how LLMs infer and apply user preferences across extended contexts. In addition to the contribution, we have considered the following possible ethical aspects of our work.

\textbf{Data Privacy and Safety.} 
\textsc{RealPref} is designed as an evaluation benchmark, and all user profiles, preferences, and interaction histories are synthetically generated, without involving real personal data. We manually reviewed and excluded ineligible data samples throughout the generation pipeline, ensuring that the generated content contains no unethical or sensitive components. 

\textbf{Dual Use.} Our work paves the way for research into smarter, user-aware LLM agents customized for individual needs, but it may also have dual-use potential. Our dataset is synthetic and provided solely for evaluation; it must not be used for unethical purposes, such as disinformation or the creation of fake profiles. As LLMs with strong preference-following capabilities move into practical deployment, we advocate for responsible, transparent deployment that prioritizes user data privacy and security, enabling users to control their data.

\textbf{Bias and Over-Personalization.} We strive to ensure preference data diversity in our dataset to minimize the chances of producing unfair or biased personalization data. Besides, our personalization tasks rely solely on preference information that users actively share with the model during dialogue. Although preferences may be expressed implicitly, the data generation is carefully designed to ensure that preference inference is grounded in clear, interpretable cues within the dialogue, rather than in speculative or unsupported assumptions, thereby reducing biased output and over-personalization. Such considerations are also required for subsequent use of the benchmark.

\textbf{Research Integrity.} Our work highlights the importance of responsible, transparent, and interpretable practices. The dataset and detailed documentation of our work, including code and prompts, are publicly available to support reproducibility. The use of AI assistants in our work is detailed in Appendix \ref{app:llm}.

\section*{Acknowledgments}
This research is supported by the Ministry of Education, Singapore, under the Academic Research Fund Tier 1 (FY2023).

\bibliography{custom}

\appendix

\section{Use of AI Assistants}\label{app:llm}
In this work, we used AI assistants for a limited scope. Regarding coding, we used AI to assist with short-form input, such as suggesting package usage and reformatting code for readability. For writing, we used AI assistants solely for the paper's language and formatting, such as grammar checking, paraphrasing, and polishing, to ensure a clear, readable delivery. All the above uses are strictly confined to the drafted content by the human authors. All AI-generated suggestions are rigorously reviewed, and all editorial decisions are reserved for human authors. AI assistants are not used for literature searches or generating text with new ideas. All the intellectual contributions presented are the work of the authors.

\section{Future Directions}
Based on the current benchmark design, there are some limitations that can be considered as directions for future research.
\begin{itemize}[itemsep=1pt,topsep=2pt,parsep=2pt]
    \item \textbf{Complex Preference and Indication}: In the real world, user preferences can change over time or depend on the condition of something else. Besides, there are other forms of preference indication, such as thumbs-up or thumbs-down for LLM responses. Enriching the dataset with such complex preferences and indications would be a possible direction.
    \item \textbf{Richer Interaction Context}: Besides simplified dialogue types, real-world interactions may involve richer, multimodal contexts. Exploring how to enrich interaction contexts is also a promising future direction.
    \item \textbf{Evaluation Extension}: The evaluation rubrics for open-ended responses could be expanded to include additional dimensions, such as fairness, privacy, and transparency, better aligning the benchmark with responsible AI principles. 
    \item \textbf{Human Subject Study}: To address the limitations of synthetic data, a longitudinal human-subject study on a real LLM platform can be considered to provide stronger evidence.
\end{itemize}

\section{Detailed Related Work}

\subsection{Personalization Benchmarks}
The user-adaptation personalization task of an LLM is to capture user preferences from historical interactions and guide subsequent response generation, ensuring the assistant's behavior aligns with user expectations without repetitive, explicit instructions. 

\textsc{PrefEval}~\cite{zhao2025llmsrecognizepreferencesevaluating} is the first benchmark to evaluate LLMs’ preference-following abilities and incorporate basic implicit preference design. However, its dialogue dataset is generated centered on a constrained set of standalone preferences, rather than on user profiles to simulate complex real-world interactions. PersonaBench~\cite{tan2025personabench} evaluates models on synthetic user profiles and private documents, but it remains a “Needle-in-a-Haystack” task focused on retrieving isolated preferences from unrelated context. Recently, interactive environments have been introduced~\cite{zhao2025personalens, qian2025userbench} to simulate real-world agent dynamics and task-oriented personalization. \textsc{PersonaMem}~\cite{jiang2025knowmerespondme} proposes a new data-generation pipeline that constructs multi-session interaction histories step by step based on dynamic user profiles, covering a variety of query tasks to test whether LLMs can retain user traits, track dynamic user preferences, and generate personalized responses in new scenarios. However, it ignores implicit preferences, and its evaluation setting does not support free-form, open-ended answers. \textsc{PersonaMem}'s constrained-generation-task evaluation requires access to the output logits of the pre-designed answers, which are not available for proprietary LLMs, and using logits is not reliable enough for evaluating free-form, open-ended answers. 

Our work draws inspiration from \textsc{PersonaMem}'s generation pipeline for multi-session dialogue generation and accounts for the different implicit nature of human preference expression, introducing a more comprehensive and granular evaluation method for genuine free-form generation tasks that is close to natural user-LLM interaction.

\subsection{Personalization Enhancing Techniques}
In general, techniques for enhancing personalization of LLMs include the following~\cite{zhangPersonalizationLargeLanguage2024}: Prompt Engineering, Retrieval-Augmented Generation (RAG), Representation Learning, and Reinforcement Learning from Human Feedback (RLHF). More recently, various Memory-Based Personalization methods have emerged.

\textbf{Prompt Engineering}~\cite{sahoo2025systematicsurveypromptengineering} is the process of designing, refining, and optimizing prompts to achieve desired outputs from language models~\cite{zhangPersonalizationLargeLanguage2024}. By incorporating user information into the prompts or providing few-shot, high-quality examples for specific tasks, LLMs can leverage contextual information and perform personalization tasks without further training.

\textbf{Retrieval-Augmented Generation}~\cite{gao2023retrieval} enhances LLM performance by retrieving relevant document segments from an external knowledge base using semantic similarity calculations~\cite{zhangPersonalizationLargeLanguage2024}. Retrieving relevant personalization information needed for the task enables LLMs to generate customized outputs.

\textbf{Representation learning} for user modeling~\cite{li2021survey, tan2023user} aims to enhance performance by learning latent representations that accurately capture each user’s behavior. This method requires a training stage, such as fine-tuning and embedding learning, which allows LLMs to adapt to specific data formats and generate responses in a particular style as instructed~\cite{zhangPersonalizationLargeLanguage2024}. This approach typically yields better results on most tasks when model parameters are available and the associated costs are acceptable.

\textbf{Reinforcement Learning from Human Feedback (RLHF)}~\cite{ouyang2022traininglanguagemodelsfollow} aims to learn an output policy based on preferential human feedback. Besides general preference human feedback, there are works on Reinforcement Learning from Personalized Human Feedback (RLPHF), such as MORL tasks~\cite{jang2023personalizedsoupspersonalizedlarge} and the P-RLHF framework~\cite{ li2024personalizedlanguagemodelingpersonalized}. By using user information as the reward signal, these methods help align LLMs with personalized preferences through reinforcement learning~\cite{zhangPersonalizationLargeLanguage2024}.

\textbf{Memory-Based Personalization} has emerged to enable an agent to exhibit a consistent personality and long-term adaptation. Recent work on persistent memory systems~\cite{zhang2025survey,hu2025memory,li2025memos} has introduced mechanisms for memory consolidation, reorganization, reinforcement, and decay. Mem0~\cite{chhikara2025mem0} proposes a scalable, persistent, and structured memory design that efficiently extracts, consolidates, and retrieves important conversational data from ongoing interactions. For personalization tasks, Persona-DB~\cite{sun2025persona} proposed an efficient database framework to address context efficiency in retrieval-augmented personalization. PRIME~\cite{zhang2025prime} integrates the cognitive dual-memory model into LLM personalization and augments it with a personalized thinking capability, mirroring episodic memory of historical user engagements and semantic memory of long-term, evolving user beliefs. PersonaVLM~\cite{nie2026personavlm} proposed hierarchical memory modules that target personalization tasks in multimodal contexts.

\section{Detailed Benchmark Design Comparison}
\subsection{Implicit Preference Comparison}
 To facilitate a better understanding of the expression types we provide in our work, we further clarify the fundamental differences between our preference expression settings and those in \textsc{PrefEval}~\cite{zhao2025llmsrecognizepreferencesevaluating}, specifically regarding what constitutes an ``implicit preference''.

\paragraph{Implicit Preferences in \textsc{PrefEval}} \textsc{PrefEval} features two types of implicit preference settings.
\begin{itemize}
    \item \textbf{Choice-based}: The LLM provides four options, and the user selects one and states preferences. This is essentially a straightforward expression of a preferred option and its reasoning. (Example: Assistant: "How about: A,... B,... C,... D,...?" User: "The documentaries don’t really interest me. How about a fun comedy or an action-packed thriller?")
    \item \textbf{Persona-driven}: A user with a brief persona explicitly reveals their preference during a discussion. (Example: "gluten-free snack due to my severe intolerance.")
\end{itemize}

\paragraph{Preference Expression in \textsc{RealPref}} Our benchmark constructs a much more complex and realistic spectrum of preference expressions based on comprehensive user profiles.
\begin{itemize}
    \item \textbf{Type 2 (Explicit Contextual Mention)}: A user discusses a topic in a single session and explicitly mentions their preference. (Note: This is the type that is relatively similar to \textsc{PrefEval}’s persona-driven setting, though with more complete profiles.)
    \item \textbf{Type 3 (Implicit Stylistic Expression)}: The user does not state preferences directly using explicit words like "like, prefer, dislike,..." Instead, they imply preferences through stylistic devices such as rhetorical questions, metaphors, contrasts, or emotional expressions. (Example: "The idea of dance classes is good! There's just something about moving to music. It is hard to understand how people drag themselves onto a treadmill every single day.")
    \item \textbf{Type 4 (Implicit Experience Feedback)}: The user develops a preference through life experiences, and this dynamic change is demonstrated across multiple sessions. (Example: Session 1: "Trying to get into a regular fitness routine... Willing to try gym workouts." -> Session 2: "I tried, but I really can't stick to the gym. I need your advice." -> Session 3: "I joined a group dance class and it feels great. This is something I can stick with.")
\end{itemize}
In conclusion, our ``implicit'' expressions are fundamentally distinct from those of \textsc{PrefEval}. We specifically emphasize Stylistic Expression (complex expression introduces retrieval challenges) and Experience Feedback (captures the dynamics of preference development across multi-session interactions).

\subsection{Evaluation Design Comparison}
Our work follows the practice in \textsc{PrefEval}~\cite{zhao2025llmsrecognizepreferencesevaluating} and further expands it to a granular scoring criterion that leverages LLM-as-a-judge to support evaluation of genuine open-ended generation tasks. While \textsc{PersonaMem}~\cite{jiang2025knowmerespondme} includes a generation task, it is actually a constrained pseudo-generation task and, functionally, a multiple-choice test: it provides the model with 4 pre-written candidate responses and checks which one the model assigns the highest probability (logits) to. It does not evaluate the model's ability to actually compose a personalized response from scratch, and its logit-matching is incompatible with proprietary LLMs (e.g., GPT-4, Claude) whose logits are inaccessible. In contrast, our work evaluates genuine free-form dialogue generation: the model is given a query and must independently retrieve the preference, plan the content, and synthesize a natural, unstructured response. We adopted the LLM-as-a-judge framework as the necessary design to enable this evaluation.

\section{Dataset Details and Examples}\label{app:dataset}
\subsection{Profiles} \label{app:profile}
\paragraph{Profile Example and Prompt.} The following is an example of a user profile. It includes a user ID, demographics information, an original persona sampled from PersonaHub~\cite{ge2025scalingsyntheticdatacreation} (its dataset card is provided below, and our usage aligns with its intention for persona-driven data synthesis), and an extended persona description. Our curated user profiles and follow-up conversations are mainly in English.
\begin{promptbox}{User Profile Example}
\begin{lstlisting}[language=json]
{
  "00001": {
    "Demographics": {
      "Name": "Alex",
      "Birth": "1999",
      "Gender": "Male",
      "Education": "Bachelor",
      "Occupation": "Music Blogger",
      "Economic Status": "Middle Class",
      "Marital Status": "Single"
    },
    "Original Persona": "a die-hard fan of Tink who runs a music review blog",
    "Extended Persona": "Alex is a passionate music enthusiast born in 1999. He's a dedicated fan of Tink and channels his love for her music into an engaging review blog. Alex, who identifies as male and is of Asian descent, brings a unique perspective to the music scene with his thoughtful analyses and compelling writing style. His blog has gained a loyal following for its authentic voice and deep dives into the nuances of music and lyrics. Outside of blogging, Alex enjoys attending live concerts and discovering up-and-coming artists in his spare time."
  }
}
\end{lstlisting}
\end{promptbox}
\begin{promptbox}{PersonaHub Dataset Card}
license: cc-by-nc-sa-4.0\\
task\_categories:\\
  - text-generation\\
  - text-classification\\
  - token-classification\\
  - fill-mask\\
  - table-question-answering\\
  - text2text-generation\\
language:\\
  - en\\
  - zh\\
tags:\\
  - synthetic\\
  - text\\
  - math\\
  - reasoning\\
  - instruction\\
  - tool\\
  - persona\\
size\_categories:\\
  - 100M<n<1B
\end{promptbox}

\begin{promptbox}{Prompt for Expanding Persona}
According to the short persona, add the following demographic information if any of them are missing: Name, Birth Year, Gender Identity, Education Level, Occupation, Economic status, Marital Status.\\

Keep the same style and pronouns, but expand the persona with additional information to around 5 sentences. You can enrich the personality, hobbies and behavioral habits if necessary to make it look real and reasonable.\\

Here is the persona: \{persona\}.\\

Return in pure JSON text format as template example: \{profile\_example\}. 
\end{promptbox}

\paragraph{Life Event Example and Prompt.} Here is an example of a user's life event, with a specific date, event description, and duration.

\begin{promptbox}{User Life Event Example}
\begin{lstlisting}[language=json]
[  
  {
    "Date": "2011/06/15",
    "Event": "Graduated with a diploma in Makeup Artistry and began working as a freelance makeup artist, taking on clients for weddings, photoshoots, and local events."
  },
  {
    "Date": "2014/03/10",
    "Event": "Collaborated with several local salons, expanding her network and gaining regular clientele, while building her professional reputation within the community."
  },
  {
    "Date": "2016/11/05",
    "Event": "Launched her social media channel dedicated to honest and detailed makeup product reviews, steadily attracting a loyal following in the beauty community."
  }
]
\end{lstlisting}
\end{promptbox}

\begin{promptbox}{Prompt for Life Event Generation}
The current version of the profile gives you the basic demographic and core persona description. \\

Here is the persona profile: \{profile\}.\\

Based on the profile, you are to expand it by generating the person's general life event history, which is composed of a list of around \{n\_events\} events.
The events are organized in json format, each event has the following keys:

- "Date": the date the event happened. (Format: year/month/day xxxx/xx/xx) 

- "Event": Event Description should be detailed and consistent with the persona. Around 1-3 sentences.

- "Duration": The event duration type. Choose from years OR months OR weeks OR days.\\

Life events refer to events that are likely to occur in this person's life. The events with duration measured by years are long-term events represent the person's milestones in his/her life. (eg. education, marriage, vocation, or other important life topics). The events with shorter duration are short-term events that are more relevant to daily life (eg. study, work, sports or other daily topics).\\

Return in pure JSON text format as template example: \{event\_templates\}.
\end{promptbox}

\subsection{Preferences} \label{app:pref}
\textbf{Generation Requirements.}
To ensure high-quality preference data in generation, we highlight requirements to ensure the generated preferences are \textit{Diverse}, \textit{Persona-Related}, \textit{Unique}, \textit{Value-Free}, and \textit{Complete}, with specific descriptions listed below:

\begin{enumerate}
\setlength{\itemsep}{0pt}
    \item \textbf{Diverse}: The preferences covering a variety of different topics and types.
    \item \textbf{Persona Related}: Each preference should be consistent with the persona's profile, sound realistic, and reflect individuality.
    \item \textbf{Unique}: Avoid generic, cliché preferences such as "dislikes fast food." Instead, provide more nuanced, distinctive preferences that are non-popular, non-obvious, and cannot be deduced by common sense.
    \item \textbf{Value-free}: Do not always align with socially desirable, universal "good", "healthy", logically "correct" preferences. Include preferences that may be unconventional or mildly negative (e.g., enjoy fast food late at night, like noisy environments). 
    \item \textbf{Complete}: The preference description should include the content and the underlying reasons, specify the type and topic.
\end{enumerate}

\textbf{Topic Pool.}
To ensure topic diversity, we introduced a topic pool comprising 7 broad categories and 37 fine-grained subtopics. The topic pool is listed in Table \ref{tab:topic_pool}.
\begin{table*}[ht]
\centering
\begin{tabular}{p{1cm}p{10cm}}
\toprule
\multicolumn{1}{c}{\textbf{Topic Fields}}   & \multicolumn{1}{c}{\textbf{Subtopics}}                                                                       \\ \midrule
\multicolumn{1}{c}{\textbf{Education}}      & Learning Style, Learning Resources                                                                                 \\ \hline
\multicolumn{1}{c}{\textbf{Entertainment}}  & Music, Movies \& Shows, Games, Reading, Online Media                                                               \\ \hline
\multicolumn{1}{c}{\textbf{Lifestyle}}      & Daily Routine, Exercise, Shopping, Beauty, Fashion, Tech Use, Home Decoration                                      \\ \hline
\multicolumn{1}{c}{\textbf{Food}}           & Cuisine, Cooking, Snacks \& Drinks, Diet, Dining Style                                                             \\ \hline
\multicolumn{1}{c}{\textbf{Travel}}         & Destinations, Activities, Companions, Pace \& Frequency, Accommodation, Transportation, Planning Style             \\ \hline
\multicolumn{1}{c}{\textbf{Work \& Living}} & Location \& Environment, Traffic \& Commuting, Climate, Work Conditions \& Hours, Work Values, Work Style \& Tasks \\ \hline
\multicolumn{1}{c}{\textbf{Social}}         & Social Energy, Friendship, Family, Romance, Social Events                                                          \\ \bottomrule
\end{tabular}
\caption{Preference Topic Pool}
\label{tab:topic_pool}
\end{table*}


\paragraph{Preference Example and Prompt.} 
The following is an example of a user preference, with an intialize event indicating how and when the user developed it. The fields include the preference and the topic, a specific date, and an event description.
\begin{promptbox}{User Preference Example}
\begin{lstlisting}[language=json]
{
  "Preference": "Prefers herbal teas and fresh fruit snacks over sugary sodas and packaged sweets, as she values a balanced lifestyle.",
  "Topic": "Food: Snacks \& Drinks",
  "Date": "2014/06/21",
  "Event": "During a beauty workshop, Emma noticed that consuming sugary sodas and snacks left her feeling sluggish, while herbal teas and fruit refreshed and sustained her energy. She made a conscious decision to adopt fresher, healthier snacks both at work and at home."
}
\end{lstlisting}
\end{promptbox}

\begin{promptbox}{Prompt for User Preference Generation}
The current version of profile gives you the basic demographic and core persona description, with several personal events.\\

Here is the persona profile: \{profile\}.\\

Based on the profile, you are to expand it by generating a list of personal preferences.

The preference should follow these types:
\{PREFERENCE\_TYPES\}.

Here are candidate preference topic pool:
\{TOPIC\_POOL\}.

The designed preference should comply with the following requirements:
\{requirements\}.\\

Step 1: Based on the profile, design a preference for each of the given topics here \{given\_topics\}.\\
Step 2: Besides the topics from Step 1, design another \{n\_add\} preferences with topics choosing from the rest topic pool that well align with the user profile.\\
You will generate \{n\_total\} preferences in total for the user.\\

Return in pure JSON text format as template example:
\{pref\_template\}.
\end{promptbox}

\begin{promptbox}{Prompt for User Preference Initialize}
The current version of profile gives you the basic demographic and core persona description, with several personal events and preference.\\

Persona profile:\{profile\}.

Personal Preference:\{preference\}.

The Preference Type Explanation: \{PREFERENCE\_TYPES\}.\\

Based on the profile, you are to expand it by generating a list of preference-related personal events.

1. Design an appropriate event for each preference, indicating that the person begins to like or dislike something after the event.

2. The event date design should be diverse and have no conflict with the personal life event.

3. The events are organized in json format, each event has following keys: \\
    - "Preference": Original given preference.\\
    - "Type": Original preference type.\\
    - "Topic": Original preference topic.\\
    - "Date": the date the event happened. (Format: year/month/day xxxx/xx/xx)\\
    - "Event": Event Description should be detailed and consistent with the persona and the related preferences. Around 1-3 sentences.\\
    
Return in pure JSON text format as template example:\{event\_templates\}.
\end{promptbox}

\paragraph{Generalized Preference Example and Prompt.} 
Generalized preferences are generated based on the user's original preferences. We utilize Few-shot Chain-of-Thought during this process to ensure that the generalized preference can be logically derived from the original ones. Below is a detailed example, including a generalized preference, referenced original preferences, and the LLM's generation rationale.
\begin{promptbox}{Generalized Preference Example}
\begin{lstlisting}[language=json]
{
  "Original Preference": [
    {
      "Topic": "Food: Cuisine",
      "Preference": "Only interested in trying global street foods or fusion cuisines, generally refusing classic fine-dining establishments which she finds unadventurous and restrictive."
    },
    {
      "Topic": "Travel: Accommodation",
      "Preference": "Prefers staying in quirky boutique hotels or artist-run guesthouses rather than large hotel chains, as the unique atmosphere inspires her work."
    },
    {
      "Topic": "Lifestyle: Beauty",
      "Preference": "Enjoys experimenting with unconventional makeup looks, such as vintage glam or fantasy-inspired themes, especially when attending local theatre events."
    }
  ]
}
{
  "Generalized Preference": {
    "Infer_reason": "Her dislike of monotonous routines, regimented systems, and classic or mainstream offerings across various domains (food, beauty, exercise, music) suggests a desire for originality and novelty.",
    "Topic": "Entertainment: Movies & Shows",
    "Preference": "Dislikes formulaic blockbuster movies and prefers independent films or experimental cinema, as unconventional storytelling stimulates her creative thinking."
  }
}
\end{lstlisting}
\end{promptbox}

\begin{promptbox}{Generalized Preference Criteria}
To ensure reasonable preference generalization, we have the following requirements:

1. \textbf{Underlying Trait}: The generalized preference must be directly derived from the same underlying reason as the original preference, without introducing new, unverified assumptions.

2. \textbf{Domain}: The inference must occur within logically adjacent domains (e.g., the same work domain: from "work emails" to "project reports") rather than leaping across unrelated domains (e.g., assuming someone who likes spicy food has an aggressive personality). Generalizations based on stereotypes or unwarranted leaps are not appropriate.\\

Example:

Known Preference: "The user strongly prefers bullet points and concise structures in work emails." -> Reasonable Generalization: "Formatting a project meeting summary using a structured outline."->Over-generalization: "Assuming the user is impatient and using blunt responses during casual chat."
\end{promptbox}

\begin{promptbox}{Prompt for Generalized Preference}
Given a list of user preferences, you are to infer several generalized user preferences based on the user's given preferences.
The new preference is different from any single given preference and has a different topic or subject.
The new preference can be rationally deduced from more than one given preferences, for the underlying reason behind the new preference being closely related with underlying reason of the given preferences.\\

For example, a user's past preference list is:

1. prefer home-cooked or artisanal meals over fast food for health benefits and quality;

2. Enjoys hiking for exercise, valuing the time spent in nature over indoor gym workouts;

3. Only uses cruelty-free beauty products, rejecting any brands that test on animals;

Then it is rational to infer that the user prefers natural, eco-friendly cleaning products at home to align with her holistic and nature-oriented lifestyle.\\

To ensure reasonable preference generalization, we have the following requirements: \{generalization\_requirements\}.

Here is the candidate preference topic pool:
\{TOPIC\_POOL\}.

Generate \{n\} new personal preferences based on the user's given preference list: \{preferences\}.

The preference contents should follow: prefer something over/dislike/aversion to another, and the underlying reasons.

Add your inference reason for the new preference, like how to deduce this new preference from the given preferences.\\

Return in pure JSON text format as template example:
\{pref\_template\}.
\end{promptbox}

\subsection{Conversations} \label{app:conv}
\subsubsection{Life-Event Conversation} \label{app:eventconv}
Life-Event Conversations indicate user profiles and life events. Through dialogue, users share background, circumstances, thoughts, and feelings. Below is an example.
\begin{promptbox}{Life-Event Conversation Example}
\textbf{User}: Hello! I've recently started working as a freelance makeup artist, and I have a wedding client next month who wants a classic vintage look. I'm a bit stuck on product choices for oily skin types. Could you recommend some foundation and powder options that hold up all day?\\
\textbf{Assistant}: Hi there! Congrats on getting a wedding client—vintage looks are always stunning. For oily skin, you'll want a mattifying, long-wear foundation and a setting powder with good oil control. Some popular options include Esté Lauder Double Wear Foundation, MAC Studio Fix Fluid, or Revlon Colorstay. As for powders, Laura Mercier Translucent Setting Powder or Rimmel Stay Matte are widely loved. Be sure to prep the skin with an oil-free primer and consider using blotting sheets for touch-ups. Would you like advice on specific application techniques for a vintage look as well?\\
\textbf{User}: Thank you, that really helps! Yes, any tips on keeping the winged liner and red lipstick looking sharp all night? Vintage looks seem easy but can be tricky under pressure. I want to make sure everything stays put for photos and dancing.\\
\textbf{Assistant}: Absolutely, vintage-inspired makeup does require a bit of extra care! For winged liner, using a waterproof gel or liquid liner will help prevent smudging—Inglot AMC Gel Liner or Stila Stay All Day Liquid Liner are good options. Set the liner with a black eyeshadow using a fine brush for added longevity. For lips, start with a lip primer, line the lips with a long-wear red liner, fill them in, then layer your lipstick (like MAC Ruby Woo or Maybelline Superstay Matte Ink), and blot between layers. Setting with a tiny bit of translucent powder through a tissue can help too. Have you been enjoying freelance work so far, or is it still new to you?\\
\textbf{User}: I actually just graduated last summer from a makeup artistry program, so this is my first full wedding season. Freelance is exciting but also a bit nerve-wracking! Thanks for all the product tips—I really want to make a great impression and hopefully get more referrals.\\
\textbf{Assistant}: That's wonderful—congratulations on launching your career! It's normal to feel a bit anxious, especially early on, but you're doing all the right things by preparing ahead. Each happy client is a stepping stone to more opportunities. If you'd ever like to chat about building your portfolio or networking with local salons, feel free to ask. Good luck with the wedding—you've got this!    
\end{promptbox}

\begin{promptbox}{Prompt for Life-Event Conversation}
You are to generate a conversation session between a user and an LLM assistant based on the user profile and history.\\
The profile gives you the basic demographic and core persona description: \{demo\}.\\
The event list gives you the user's personal history event:
\{events[:i]\}.\\
In the conversation, the user is interacting with LLM assistant between \{time\_range\}.\\
In the time range, the followeing life event have already happend: \{prev\_event\}. And the following life event is upcoming: \{nxt\_event\}.\\

Generate \{n\_sessions\} multi-turn conversations (2-4 rounds each) where the user is at the life stage described above.

1. Each session should begin with the User initiating the conversation, guided by one of the following dialogue intents:\\
- Task-based request: asking for specific help, knowledge, resources, advice, or content creation.\\
- Life choice discussion: exploring academic, career, or future path options.\\
- Emotional companionship: expressing feelings such as confusion, anxiety, frustration, excitement, or achievement.\\
2. Throughout the dialogue, let the User gradually and naturally reveal aspects of their background, life circumstances, thoughts, and emotions.\\
3. The Assistant should start with no knowledge of the user's persona or situation. All personal details must be disclosed progressively by the User during the conversation.\\
4. The Assistant's responses should be insightful, detailed, and constructive—providing useful suggestions, gentle reasoning, or professional guidance. Whenever possible, offer multiple perspectives, options, or explanations. Keep the tone friendly, conversational, and engaging, rather than overly formal.\\
5. Ensure the dialogues feel realistic and natural:\\
- Vary word choice, sentence structures, and tones across sessions.\\
- Keep content coherent and contextually appropriate.\\
- Make each conversation sufficiently complete and meaningful.\\

Return in pure JSON text format as template example:
\{conversation\_template\}.
\end{promptbox}

\subsubsection{Random Conversation} \label{app:ranconv}
Random Conversations are about common-sense questions and will not reveal user-specific information. Below is an example with the assistant response omitted for simplicity of demonstration.

\begin{promptbox}{Random Conversation Example}
\textbf{User}: I'm having trouble connecting to Wi-Fi on Windows 10. Any troubleshooting tips?\\
\textbf{Assistant}: [Response]\\
\textbf{User}: What if none of these suggestions work?\\
\textbf{Assistant}: [Response]
\end{promptbox}

\begin{promptbox}{Prompt for Random Conversation}
Generate {n} sessions of multi-turn conversations (2-5 rounds each) between a user and a warm, articulate, and helpful LLM assistant.\\

1. Each session should avoid topics involving personal background details, life achievements, or user personal preferences.\\
2. The user initiates the session with a realistic, detailed question or request reflect real-world use cases such as:\\
    - asking factual questions or explanations (e.g., science, technology, society)\\
    - troubleshooting or tool usage (e.g., Python, Excel, Git, API, etc.)\\
    - general curiosity or reasoning questions\\
    - simple small talk or clarification requests\\
    - text generation tasks\\
3. Vary the topics and tones to make the dataset diverse. You can choose topics randomly from domains such as technology, science, education, programming, history, art, and productivity tools. \\
4. The assistant should provide thoughtful, detailed, and constructive replies, offering multiple ideas, explanations, or suggestions where appropriate. The tone should remain friendly and engaging, never overly formal.\\
5. Ensure the conversation session feels contextually complete.\\

Return in pure JSON text format as template example:
\{conversation\_template\}\\
\end{promptbox}

\subsubsection{Preference-Expression Conversation} \label{app:prefconv}
Preference-Expression Conversations indicate user preference. We provide four \textit{Preference Expression Types} from explicit to implicit, each with an example listed below.

\paragraph{1. Explicit - Direct Statement} Preferences are stated directly and explicitly in a single turn without redundant words.
\begin{promptbox}{Direct Statement Example}
\textbf{User}: Hi! I want to get more physically active, but \textbf{I really avoid routine gym workouts because they're monotonous and uninspiring for me. I much prefer spontaneous activities like dancing at local community events.} Any ideas for keeping fit that aren't boring gym routines?\\
\textbf{Assistant}: Absolutely, I understand that traditional gym workouts aren't for everyone—it's important to have fun while staying active! Here are a few ideas that can keep things lively and spontaneous:  1. Dance Classes or Social Dancing: Try drop-in classes for salsa, swing, or even hip-hop at local studios or community centers. Social dance nights are great for meeting people and get your heart rate up without feeling like exercise.  2. Outdoor Group Activities: Join recreational sports leagues (like volleyball, ultimate frisbee, or even kickball) where the focus is on fun and socializing rather than intense competition.  3. Hiking Groups or Urban Walks: Explore nature trails, city walking tours, or even organized hiking meet-ups. Each outing is a little different and can double as a mini-adventure.  4. Fitness Meetups: Check out community boards or social media for pop-up yoga in the park, outdoor bootcamps, or even roller skating meetups.  5. Volunteering: Since you enjoy community events, active volunteering (like setting up events, ushering at theaters, or organizing local festivals) can keep you moving while contributing to a good cause.  Mixing things up and following what excites you can make staying active something you actually look forward to. Would you like more suggestions tailored to your interests or info on how to find local events?
\end{promptbox}

\begin{promptbox}{Prompt for Direct Statement}
You are to generate 1 session of conversation between a user and an llm assistant based on the user profile and user preference.\\
The profile gives you the basic demographic and core persona description. Here is the persona profile: \{profile\}.\\
The preference gives you the preference content that the user should indicate in this conversation. Here is the preference: \{pref\}.\\
The conversation occurs between time range \{time\_range\}.\\
The event list gives you the user's personal life event history: \{events\}.\\

Instruction:\\
- In the conversation, the user stated the preference directly, CLEARLY and EXPLICITLY in a single turn without redundant words, and the assistant gives a reponse.\\
- The LLM assistant in the conversation DO NOT KNOW anything about the user, all such personal information should be revealed by user in the conversation, LLM assistant just work as a general AI assistant for question answering.\\
- The LLM assistant gives insightful, detailed, and constructive responses, offering useful suggestions, gentle reasoning, or professional guidance. Let the assistant provide more than one idea or explanation when possible. Keep the tone friendly and engaging, not overly formal.\\
- Remember to keep the dialogue realistic, with diversity in word, sentence structures and tone, and the conversation content reasonable and complete.\\

Return in pure JSON text format as template example:
\{conversation\_template\}.\\
\end{promptbox}

\paragraph{2. Explicit - Contextualized Mention} In multi-round conversations, users start the session with other related daily topics as context instead of directly stating preferences. As the dialogue advances, preferences are naturally mentioned explicitly in the context.
\begin{promptbox}{Contextualized Mention Example}
\textbf{User}: Hi there! I was looking for some ideas to keep my weekends fun and active without spending too much money. Got any suggestions?\\
\textbf{Assistant}: Absolutely! There are plenty of ways to have an active and enjoyable weekend without breaking the bank. Here are a few ideas:  1. Go for a nature walk or hike in a local park. 2. Join a community sports group that has free or low-cost sessions. 3. Try a YouTube dance or workout class at home with friends. 4. Attend local events or street fairs, which often offer free activities and entertainment. 5. Organize a picnic with friends and some outdoor games like frisbee or badminton.  Would you like more suggestions tailored to your interests or any particular type of activity?\\
\textbf{User}: Thanks! I like the idea of attending local events. Lately, I've noticed several community dance nights popping up nearby. \textbf{Honestly, I've never been into routine gym workouts—they just feel so monotonous and uninspiring to me. I much prefer spontaneous activities like dancing. Do you think joining community dance events counts as good physical activity?} Any tips for someone just jumping in?\\
\textbf{Assistant}: Dancing at community events absolutely counts as excellent physical activity! It's fantastic for your cardiovascular health, flexibility, balance, and mood. Plus, the energy and social aspect make it much more enjoyable than routine workouts for many people.  Here are some tips for getting started:  1. Don't worry about technical skills—community events are usually very inclusive and welcoming to beginners. 2. Wear comfortable shoes and clothes so you can move freely. 3. Stay hydrated, as dancing can be quite a workout. 4. Bring a friend along if you feel shy at first, or just go with an open mind to meet new people. 5. Try out different dance styles to see what you enjoy the most—salsa, swing, folk, or even freestyle.  And the best thing is, you're having fun, so the exercise doesn't even feel like a chore! Would you like some resources to look up local dance nights or maybe music playlists for practice at home?\\
\textbf{User}: That's so reassuring! I definitely vibe with the fun, social energy rather than repetitive gym routines. I'd love some tips for finding more local dance nights or any online groups where people organize spontaneous dance meetups. Also, if you have music playlist suggestions for home practice, I'm all ears!\\
\textbf{Assistant}: Great to hear you're excited! For finding local dance nights:  1. Check out community center bulletin boards or websites—they often list weekly or monthly events. 2. Browse social media platforms (like Facebook Events or Meetup.com) using keywords like “community dance,” “social dance,” or specific styles (e.g., “salsa night,” “swing party”). 3. Some local coffee shops or cultural centers host themed dance nights—ask around next time you're out! 4. Chat with local dance studios; even if you don't want structured classes, they often know about informal gatherings.  For online groups, Meetup.com and Facebook Groups are especially helpful. Try searching for your city + “dance” or “jam session.”  Music playlists: There are great ones on Spotify and YouTube for various moods—check out playlists titled “Dance Party Classics,” “Salsa \& Bachata Hits,” or “Feel-Good Indie Dance.”  Let me know if you'd like me to curate a sample playlist or help you draft a message to join an online group!
\end{promptbox}

\begin{promptbox}{Prompt for Contextualized Mention}
You are to generate 1 session of conversation between a user and an llm assistant based on the user profile and user preference.\\
The profile gives you the basic demographic and core persona description. Here is the persona profile: \{profile\}.\\
The preference gives you the preference content that the user should indicate in this conversation. Here is the preference: \{pref\}.\\
The conversation occurs between time range \{time\_range\}.\\
The event list gives you the user's personal life event history: \{events\}.\\

Instruction:\\
1. The conversation has >= 3 rounds.\\
2. In the conversation, the user will NOT directly talking about the preference, but start a conversation with some other related daily life topics or request as context.\\
3. As the conversation progresses, the user later NATURALLY mention the preference CLEARLY and EXPLICITLY in contextual conversations.\\

- The LLM assistant in the conversation DO NOT KNOW anything about the user, all such personal information should be revealed by user in the conversation, LLM assistant just work as a general AI assistant for question answering.\\
- The LLM assistant gives insightful, detailed, and constructive responses, offering useful suggestions, gentle reasoning, or professional guidance. Let the assistant provide more than one idea or explanation when possible. Keep the tone friendly and engaging, not overly formal.\\
- Remember to keep the dialogue realistic, with diversity in word, sentence structures and tone, and the conversation content reasonable and complete.\\

Return in pure JSON text format as template example:
\{conversation\_template\}.\\    
\end{promptbox}

\paragraph{3. Implicit - Stylistic Expression} The context resembles the second type, but users will not state preferences with explicit words like Like, Prefer, or Dislike. Instead, they imply preferences through stylistic devices such as rhetorical questions, metaphors, contrasts, or emotional expression.
\begin{promptbox}{Stylistic Expression Example}
\textbf{User}: Hey there! I've been thinking about switching up my daily routine lately. Got any creative ways to recharge after a long day at work? My schedule's been packed with makeup clients, and I could use something to shake off the stress.\\
\textbf{Assistant}: Absolutely! After a busy day, it's great to have unwinding rituals that actually excite you. Some options could be a little home spa session—think an aromatic bath, face mask, and soothing playlist—or even picking up a fun hobby like painting or baking. If you prefer something active, you could try a walk in a nearby park, cycling, or maybe signing up for a local art or dance class. Do any of those sound appealing, or are there activities you used to enjoy that you haven't revisited in a while?\\
\textbf{User}: The idea of dance classes is good! There's just something about moving to music that makes me forget the clock even exists. Honestly, I've never understood how people drag themselves onto a treadmill every single day—just the thought of it puts me to sleep. I keep stumbling into little community salsa nights or street performances after work lately, and those improvised jams end up being the highlight of my week.\\
\textbf{Assistant}: Dance is such a fantastic way to reinvigorate your body and mind, especially with the social and creative atmosphere of community events. Unlike gym routines, which can become a bit mechanical, dancing taps into expression and connection—plus, it's a great workout! If you love these spontaneous dance nights, you might also enjoy joining a local dance group or checking for upcoming themed events in your neighborhood. You could even invite a few friends or fellow artists along for extra fun! And if you ever want tips for stretching or caring for your body post-dance, I'm here to help.\\
\textbf{User}: Exactly! I guess it makes sense—I'd rather end up in a train of laughter trying to follow someone else's footwork than counting reps in a gym mirror. I get my fill of routine at work; when I'm off the clock, it's about freedom and a fresh vibe. Sometimes I mix it up with some vintage shopping at the weekend markets too, especially if there's live music. Life's too short not to move with the rhythm, right?\\
\textbf{Assistant}: That's a wonderful philosophy! Blending movement, music, and even a bit of vintage treasure hunting sounds both energizing and inspiring. These experiences not only help you unwind but probably fuel your creativity for your makeup artistry as well. If you ever want unique ideas to incorporate your passions—like organizing a mini makeover session at a dance event or styling vintage market-goers—let me know. Embracing spontaneity is truly an art in itself!
\end{promptbox}

\begin{promptbox}{Prompt for Stylistic Expression}
You are to generate 1 session of conversation between a user and an llm assistant based on the user profile and user preference.\\
The profile gives you the basic demographic and core persona description. Here is the persona profile: \{profile\}.\\
The preference gives you the preference content that the user should indicate in this conversation. Here is the preference: \{pref\}.\\
The conversation occurs between time range \{time\_range\}.\\
The event list gives you the user's personal life event history: \{events\}.\\

Instruction:\\
1. The conversation has >= 3 rounds.\\
2. In the conversation, the user will NOT directly talking about the preference, but start a conversation with some other related daily life topics or request.\\
3. As the conversation progresses, the user later NATURALLY mention the preference in contextual conversations.\\
4. The user will NOT directly express preference by explicit words such as LIKE, PREFFER, or DISLIKE. Instead, user IMPLICITLY indicate the preference by rhetorical questions, metaphors, contrasts, or emotional expression. For example, to express 'I prefer science fiction movies to art films', one can say \\
---- 'I dont get how people stay awake through movies like The Tree of Life.'\\
---- 'I tried watching The Tree of Life, but honestly, I'd much rather rewatch Dune.'\\
---- 'I've watched Blade Runner 2049 like five times. Never gets old.'\\
---- 'Dune just hits different. So immersive.'\\

- The LLM assistant in the conversation DO NOT KNOW anything about the user, all such personal information should be revealed by user in the conversation, LLM assistant just work as a general AI assistant for question answering.\\
- The LLM assistant gives insightful, detailed, and constructive responses, offering useful suggestions, gentle reasoning, or professional guidance. Let the assistant provide more than one idea or explanation when possible. Keep the tone friendly and engaging, not overly formal.\\
- Remember to keep the dialogue realistic, with diversity in word, sentence structures and tone, and the conversation content reasonable and complete.\\

Return in pure JSON text format as template example:
\{conversation\_template\}.\\
\end{promptbox}

\paragraph{4. Implicit - Experience Feedback} Multiple sessions, based on the user preference Initialize Event. The user won’t explicitly state their preference in one session but will reveal it gradually through experience feedback across sessions. 
\begin{promptbox}{Experience Feedback Example}
\textbf{[Session 1]}\\
\textbf{User}: Hey! I’m thinking of trying to get into a regular fitness routine. I always see people raving about the benefits of gym workouts, but I’ve never really stuck with any for long. Do you have any tips for someone who gets bored easily? I want to keep myself motivated!\\
\textbf{Assistant}: Hi! That’s a great initiative. Keeping fitness fun and engaging can be a real challenge, especially if you find traditional routines monotonous. Here are a few suggestions you might find helpful:  1. Mix up your activities: Try incorporating different kinds of exercise like swimming, hiking, cycling, or group classes. This helps keep things fresh and less predictable. 2. Try out classes: Many gyms and studios offer dance, kickboxing, or yoga classes—you can sample various ones to see what feels most enjoyable. 3. Set mini-goals: Instead of focusing on a single workout type, try setting goals related to new skills or experiences—like mastering a yoga pose or running a local fun run. 4. Get a workout buddy: Having a friend to join you can boost motivation and add a social element to your routine.  Let me know if you want more specific ideas, or if any of these sound interesting to you!\\
\textbf{User}: Thanks! I like the idea of mixing things up. I haven’t tried any group classes yet, but maybe I should. I do find myself zoning out doing the same treadmill or weights routine over and over, so something a bit more dynamic could definitely help.\\
\textbf{Assistant}: Absolutely! Group classes often have an energetic vibe that can make the whole experience more enjoyable and less repetitive. Dance-based fitness classes like Zumba or hip-hop, or even things like spinning with music, can keep you engaged. If you’re not into classes, you could also look out for local community events like fun runs, charity walks, or weekend sports meetups. Sometimes being in a new environment with others doing something active can break the monotony, too. And if you ever need advice on balancing your fitness routine with a busy work schedule, just let me know!\\

\textbf{[Session 2]}\\
\textbf{User}: Hey again! I wanted to share—last weekend I did something totally out of my comfort zone and joined a flash mob dance group for a community event. I honestly signed up on a whim, but it turned out to be so much more fun than any gym session I’ve had.\\
\textbf{Assistant}: That sounds amazing! Flash mobs can be such an exhilarating experience—not only do you get a fantastic workout, but you’re part of something creative and vibrant with others. Dancing is a great way to get exercise without it feeling like a chore. What was it about the event that you enjoyed most—the music, the group energy, or just the spontaneity?  If you’re looking to keep the momentum going, you might want to look for similar events or dance meetups in your area. There are also lots of dance fitness classes, or you could organize your own mini dance sessions with friends! Keeping activities enjoyable and social is a great way to build exercise into your lifestyle without it ever feeling monotonous.\\
\textbf{User}: Honestly, it was all of it—the music, being with a group, and just letting go and moving. I felt so energized afterwards. It didn’t feel like a workout at all! I wish more fitness experiences could be like that. I think I might look out for other dance or movement-based things to do in the city.\\
\textbf{Assistant}: I love hearing that! When exercise feels joyful and connected, it’s so much easier to stay consistent—and you end up looking forward to it instead of just checking it off a list. Dance, in particular, combines movement, creativity, and social interaction, which can all boost your mood and motivation. Keep an eye out for local dance events, street festivals with open dance floors, or even themed fitness parties. And don’t overlook opportunities like community theatre—volunteering to help with performance prep or being involved backstage can be surprisingly active, too!  If you want suggestions for specific events to try or tips on discovering local options, I’d be glad to help.
\end{promptbox}

\begin{promptbox}{Prompt for Experience Feedback}
You are to generate \{n\_sessions\} sessions of conversation between a user and an llm assistant based on the user profile and user preference.
The profile gives you the basic demographic and core persona description.\\
Here is the persona profile: \{profile\}.\\
The preference gives you the preference content that the user should indicate in this conversation. Here is the preference: \{pref\}.\\
This preference stems from the following past event: "\{pref\_event\}" happened on \{start\_date\}.\\
The event list gives you the user's personal life event history: \{events\}.\\

Instruction:\\
1. Each session has 1-3 rounds of conversation. \\
2. In the conversation, the user will NOT directly talking about the preference, but start a conversation with some other related daily life topics or request.\\
3. The session content stems from the past event. You should further enrich the personal experience story by including details or extra information in preference event.\\
4. The conversation timestamp is related to the date of past event (\{start\_date\}), keep the timeline consistent.\\
5. Do not directly state the preference or event experience in one session, and the user will NOT directly express preference by explicit words such as LIKE, PREFFER, or DISLIKE. Instead, the full information should gradually emerge across 2 to 3 sessions of user-llm interactions, indicating by fragmented personal experiences, behaviour and casual remarks. For example,\\
    - The early session may happen before the event, when the user has not yet developed such a preference, and is willing to try.\\
    - The later sessions may happen after the event experience, where the user may share event experience, give feedbacks and indicate preference across session interactions.\\
6. Key Reminder: The assistant cannot infer the user's preferences from any single session dialogue. It can only infer the user's preferences by integrating the dialogue information of all sessions.\\
7. The assistant should always reply with helpful, warm, and constructive responses, but DO NOT infer the full preference too early.  \\

- The LLM assistant in the conversation DO NOT KNOW anything about the user, all such personal information should be revealed by user in the conversation, LLM assistant just work as a general AI assistant for question answering.\\
- The LLM assistant gives insightful, detailed, and constructive responses, offering useful suggestions, gentle reasoning, or professional guidance. Let the assistant provide more than one idea or explanation when possible. Keep the tone friendly and engaging, not overly formal.\\
- Remember to keep the dialogue realistic, with diversity in word, sentence structures and tone, and the conversation content reasonable and complete.\\

Return in pure JSON text format as template example:
\{conversation\_template\}.\\
\end{promptbox}

\section{Evaluation Details}\label{app:eval}
\subsection{Preference Query} \label{app:query}
Queries are designed based on the target preferences. High-quality queries should ensure that, without knowing the preferences, \textbf{answering the question naturally is likely to violate them}~\cite{zhao2025llmsrecognizepreferencesevaluating}. Therefore, the query content should not reveal user preferences or any underlying user intention. Here are examples to facilitate understanding:
\begin{promptbox}{Preference Query Example Analysis}
\textbf{Example 1:}\\
\textbf{Preference}: I prefer companies that allow unlimited sick days.\\
\textbf{Bad Query}: Which companies have the most generous sick leave policies?\\
\textbf{Analysis}: The query directly reveals what the preference is looking for.\\
\textbf{Good Query}: I'm moving to Seattle and seeking job opportunities. Can you recommend a suitable company?\\
\textbf{Analysis}: The query did not reveal any user information regarding company preferences.\\

\textbf{Example 2:}\\
\textbf{Preference}: I prefer staying in traditional accommodations over modern hotels to experience cultural authenticity.\\
\textbf{Bad Query}: I'm planning a vacation to fully immerse myself in the local culture. Can you recommend accommodation for me?\\
\textbf{Analysis}: A bad example. The query directly leaked the user's intention.\\
\textbf{Good Query}: I'm planning a vacation to Japan. Can you recommend accommodation for me?\\
\textbf{Analysis}: The query contains only the core question without leaking the user's preference and intention.
\end{promptbox}

The following is a complete query example. The fields include the target preference, an open-ended question with an explanation, 4 MCQ options with the correct answer labeled, and 2 true-or-false questions derived from the MCQ options.
\begin{promptbox}{Preference Query Example 1}
\begin{lstlisting}[language=json]
{
  "Preference": "Prefers staying in traditional accommodations, such as ryokans, over modern hotels to experience cultural authenticity and personalized service.",
  "Query": "I'm looking for a place to stay during my trip to Japan. What kind of accommodation would be best for me?",
  "Explanation": "The expected response should highlight traditional aspects associated with cultural authenticity, aligning with the user's desire for personalized service and cultural immersion.",
  "MCQ_options": {
    "A": "A sleek hotel in downtown Tokyo with smart tech and 24/7 room service.",
    "B": "A traditional ryokan in Kyoto that serves kaiseki meals and offers onsen baths.",
    "C": "A capsule hotel near Shinjuku station that's convenient and budget-friendly.",
    "D": "A modern Western-style hotel with a rooftop bar and fitness center."
  },
  "Correct_Answer": "B",
  "T_Query": "I'm looking for a place to stay during my trip to Japan. Do you think a traditional ryokan in Kyoto that serves kaiseki meals and offers onsen baths would be the one for me?",
  "F_Query": "I'm looking for a place to stay during my trip to Japan. Do you think a modern Western-style hotel with a rooftop bar and fitness center would be the one for me?"
}
\end{lstlisting}
\end{promptbox}

Regarding the comparison of MCQ and True-or-False questions, here is an example query of a negative preference (dislike or aversion to something) to facilitate understanding. In the 4 MCQ options below, 3 incorrect options share the same characteristic — routine workouts —while the correct one obviously stands out and can be guessed even if the model has no knowledge of use preference. In contrast, the true-or-false format avoids this possibility, and the model must decide based on the current single option.
\begin{promptbox}{Preference Query Example 2}
\begin{lstlisting}[language=json]
{
  "Preference": "Avoids routine gym workouts because she finds them monotonous and uninspiring, preferring spontaneous physical activities like dancing at community events.",
  "Query": "I'm interested in finding ways to stay active in my free time, what would you suggest?",
  "Explanation": "Without knowing the preference, an assistant is likely to recommend standard gym workouts or classes, which can easily violate the user's aversion to monotonous exercise routines. The query is sufficiently broad, making it challenging to guess the user's specific dislike for routine workouts versus other forms of activity.",
  "MCQ_options": {
    "A": "Join local community dance events and try different styles whenever you feel like it.",
    "B": "Take group fitness classes at a nearby gym several nights a week.",
    "C": "Follow an at-home structured workout program with daily routines.",
    "D": "Plan a weightlifting schedule and stick to it for consistency."
  },
  "Correct_Answer": "A",
  "T_Query": "I'm interested in finding ways to stay active in my free time, what would you suggest? Do you think joining local community dance events and trying different styles whenever you feel like it would be the one for me?",
  "F_Query": "I'm interested in finding ways to stay active in my free time, what would you suggest? Do you think following an at-home structured workout program with daily routines would be the one for me?"
}
\end{lstlisting}
\end{promptbox}

\begin{promptbox}{Prompt for Preference Query}
Given a user preference, you are to generate an open-ended query, corresponding multiple-choice question (MCQ) options, and True or False query designed to test whether a language model assistant understands the user's preference.\\

Instruction:\\
1. Open-ended query\\
    (1) The query should be a natural-sounding open-ended question, concise in 1 sentence, written from user's first-person perspective.\\
    (2) The query is designed based on user preference, requiring that without knowing the preference, naturally answering the question can easily violate the user's preference. Here are bad and good examples for your information.\\
        - Bad example: \{bad\_example\}.\\
        - Good example: \{good\_example\}.\\
    (3) The query should be concise. DO NOT include anything about the user preference or underlying user purpose or intention in the query. Here are bad and good query examples toward one preference for your information.\\
        - Preference: \{example\_preference\}.\\
        - Bad example: \{bad\_example\}.\\
        - Good example: \{good\_example\}.\\
    (4) DO NOT generate contradictary or obvious question such that the question either directly contradicts the user's preference or is so aligned that it is not challenging.\\

2. Multiple-choice question\\
    (1) For Open-ended query, design four possible options, each is concise in 1 sentence. \\
    (2) All four options are perfectly relevant, reasonable and common sense answers to the query. None of them contain obvious common sense errors or logical errors.\\
    (3) The four MCQ options has ONLY ONE answer that BEST aligns with the user's preference.\\
    (4) Make sure that the correct answer can not be inferred purely based on common sense. Only the one aware of the user preference can figure out the correct option.\\
    (5) DO NOT explain, repeat anything about the user preference or underlying user purpose in any options. Do not add extra explanation about the options in MCQ.\\

3. True or False\\
    (1) The True or False query is derived from the open-ended query and the MCQ options: [open-ended query] + Do you think [option] would be the one for me?\\
    (2) T\_Query simply combines the query and the correct option from MCQ, and F\_Query simply combines the query and one of the false option from MCQ. \\

Key Reminder: Add a explanation about why it is challenging in answering the query without violate the preference. The key is, without knowing the preference, the assistant is highly likely to violate it when answering the question, which means suggesting options that is contrary to the user's preference.\\

Here is the user preference: \{pref\}.\\

Return in pure JSON text format as template example:
\{qa\_template\}.\\
\end{promptbox}

\subsection{Evaluation Rubrics} \label{app:rubrics}
The scoring criteria for open-ended answers in this work have 3 rating dimensions described below. In practice, the judge LLM will be provided with a preference, a query, and the rubrics for the specified dimension, and will be prompted to assign a score and provide a judge's rationale.

\textbf{Preference Awareness}: focuses on the extent to which the answers recognize or understand user preferences. See detailed rubrics in Table \ref{tab:rub1}.

\textbf{Preference Alignment}: focuses on how well the answer content aligns with the user's preferences. See detailed rubrics in Table \ref{tab:rub2}.

\textbf{Answer Quality}: focuses on the extent to which the answer content is actually helpful and constructive to the user. See detailed rubrics in Table \ref{tab:rub3}.

\begin{promptbox}{Prompt for LLM-as-a-Judge Evaluation}
You are a strict evaluator. Given a user preference, a query and a candidate assistant response, score the response on a 1-5 scale, where 1 = very poor and 5 = excellent. Here is the evaluation rubric: \\

\{evaluation rubric\}\\

User preference: \{preference\}.\\
User query: \{query\}.\\
Assistant response: \{response\}.\\

Your judge should be organized in JSON format with the following 2 keys: \\
- "Judge\_Reason": A very short judge reason how you evaluate the response according to the rubrics.\\
- "Score": The score you give to the response.\\

Return in pure JSON text format as template example:
\{judge\_template\}.\\
\end{promptbox}

\begin{table*}[htbp]
\centering
\begin{normalsize}
\begin{tabular}{p{1cm}p{13cm}}
\toprule
\multicolumn{1}{c}{\textbf{Score}}   & \multicolumn{1}{c}{\textbf{Criteria}}                                                                       \\ \midrule
\multicolumn{1}{c}{\textbf{1}}      & No mention or reflection of user preference at all.                                                                                 \\ \hline
\multicolumn{1}{c}{\textbf{2}}  & Mentions user preference, but the stated preference content is incomplete or incorrect.\\ \hline
\multicolumn{1}{c}{\textbf{3}}      & Mentions user preference, and correctly states the preference, but not well integrated in the response (the response is highly generic, which can be applied to any user). \\ \hline
\multicolumn{1}{c}{\textbf{4}}           & Mentions user preference, correctly states the preference, and integrates the preference in the response.\\ \hline
\multicolumn{1}{c}{\textbf{5}}         & Mentions user preference, correctly states the preference, and the response is fully aware, deeply focused, consistently centered on the preference.\\ \bottomrule
\end{tabular}
\caption{Evaluation Rubrics for Preference Awareness}
\label{tab:rub1}
\end{normalsize}
\end{table*}

\begin{table*}[ht]
\centering
\begin{normalsize}
\begin{tabular}{p{1cm}p{13cm}}
\toprule
\textbf{Score} & \multicolumn{1}{c}{\textbf{Criteria}}                                                                     \\ \midrule
\multicolumn{1}{c}{\textbf{1}}     & \begin{tabular}{p{13cm}}\textbf{Clear violation of user preference.} Examples:\\ - User prefers A over B, but the response only suggests B.\\ - User only likes A, but the response only suggest something except A.\\ - User does not accept B, but the response only suggest B.\end{tabular}                                                    \\ \hline
\multicolumn{1}{c}{\textbf{2}}     & \begin{tabular}{p{13cm}}\textbf{Mostly inconsistent or ignores some taboos.} Examples:\\ - User only likes A, but the response suggest A and something except A.\\ - User does not accept B, but the response also contains B.\end{tabular}                                                                                                \\ \hline
\multicolumn{1}{c}{\textbf{3}}     & \begin{tabular}{p{13cm}}\textbf{Generally aligned but with inaccuracies or gaps.} Examples:\\ - User prefers A over B, but the response suggests A as well as B.\\ - User likes A, but the response only suggest something other than A.\end{tabular}                                                                                                                                 \\ \hline
\multicolumn{1}{c}{\textbf{4}}     & \begin{tabular}{p{13cm}}\textbf{Aligned and mostly consistent.} Examples:\\ - User prefers A over B, and the response suggests A.\\ - User only likes A, and the response suggests A.\\ - User does not accept B, and the response does not contain B.\\ - User likes A, and the response is mostly about A.\end{tabular} \\ \hline
\multicolumn{1}{c}{\textbf{5}}     & \begin{tabular}{p{13cm}}\textbf{Fully aligned, no conflicts at all} (requirement for score 4), and the response is consistently centered on the preference, with \textbf{tailored suggestions and explanations.} \end{tabular}    \\ \bottomrule                                                                        
\end{tabular}
\caption{Evaluation Rubrics for Preference Alignment}
\label{tab:rub2}
\end{normalsize}
\end{table*}

\begin{table*}[h]
\centering
\begin{normalsize}
\begin{tabular}{p{1cm}p{13cm}}
\toprule
\textbf{Score} & \multicolumn{1}{c}{\textbf{Criteria}}                                                                     \\ \midrule
\multicolumn{1}{c}{\textbf{1}}     & \begin{tabular}{p{13cm}}\textbf{Off-topic or irrelevant.} The response does not address the question, only irrelevant information.\end{tabular}                                                    \\ \hline
\multicolumn{1}{c}{\textbf{2}}     & \begin{tabular}{p{13cm}}\textbf{On-topic but no useful information.} Example:\\
        - The response is not informative and seems like background information or small talk rather than actual advice.\\
        - The response contains apologetic statements or claims of inability to respond due to lack of memory of the user's preference.\\
        - The assistant asks for more information without providing any substantive suggestions, and advises directly related to the query.\end{tabular}                                                                                                \\ \hline
\multicolumn{1}{c}{\textbf{3}}     & \begin{tabular}{p{13cm}}\textbf{On-topic, some relevant but not constructive information.} Example:\\
        - The response only gives some abstract solutions; the advice is vague and unactionable.\\
        - The response provides some relevant recommendations, but also contains apologetic statements or claims of inability to follow the user's preference.\end{tabular}                                                                                                                                 \\ \hline
\multicolumn{1}{c}{\textbf{4}}     & \begin{tabular}{p{13cm}}\textbf{Constructive, useful and gives clear guidance.} The response provides specific, relevant recommendations or advice related to the query without any apologetic statements or claims of inability to respond.\end{tabular} \\ \hline
\multicolumn{1}{c}{\textbf{5}}     & \begin{tabular}{p{13cm}}\textbf{Highly constructive, detailed, multiple helpful ideas or perspectives.} \end{tabular}    \\ \bottomrule                                                                
\end{tabular}
\caption{Evaluation Rubrics for Answer Quality}
\label{tab:rub3}
\end{normalsize}
\end{table*}

\section{Benchmark Quality Verification}\label{app:quality}
To ensure the quality of our benchmark work, we have conducted both automated and manual verification of each benchmark component, including the synthetic data, the queries, and the evaluation scoring criteria. Human annotators (graduate students who are contributors to this work) are included in this part. They understand and consent to the subsequent use of the collected data.
\subsection{Synthetic Data}
Real user data (preferences, profiles, conversations) is highly private and user-specific, and user behavior is complex, making it challenging to bridge the gap between synthetic categories and real users' long-term behavior. Our work aims to move a step closer to natural user-LLM dialogue by using diverse, well-constructed synthetic data. We have always recognized the importance of data quality and have made it central to our data generation process.

\paragraph{Profiles and Preferences} 
Following our generation pipeline, we performed manual checks and filtering early in the process of creating user profiles and preferences. 
\begin{enumerate}
    \item \textbf{Persona}: We randomly sampled 150 candidate personas from PersonaHub~\cite{ge2025scalingsyntheticdatacreation}, a massive, open-source dataset comprising 1 billion diverse personas, each represented as a brief sentence. We conducted the first manual check and filtering of these personas, deleting the rare or unnatural ones (e.g., those describing a couple rather than a person, or those related to inappropriate backgrounds such as WWII or Paranormal Events), leaving 137 personas for the next step.
    \item \textbf{Profile}: The generated persona set is then used to construct a detailed profile that includes basic demographics and an extended persona description. We perform a second manual check on the plausibility of generated profiles and retain 100 profiles with diverse, appropriate and plausible attributes.
    \item \textbf{Life Events and Preferences}: Based on the resulting profiles, we generated 12 personal life events and 12 preferences for each user. These events and preferences are then subjected to manual and LLM-assisted plausibility checks, and 10 preferences and 8-10 events are retained per user.
    \item \textbf{Generalized Preferences}: Generalized preferences are based on the original preferences. During generation, we manually checked whether each generalized preference can be rationally inferred from the original preferences. The method is similar to the "Easy Test" we provided in Appendix~\ref{app:infer}: giving all original preferences and choosing which option may align with the user's preference without viewing the ground-truth generalized preference. The validated samples remain in the dataset. 
\end{enumerate}

\begin{figure*}[t]
    \centering
    \includegraphics[width=0.9\linewidth]{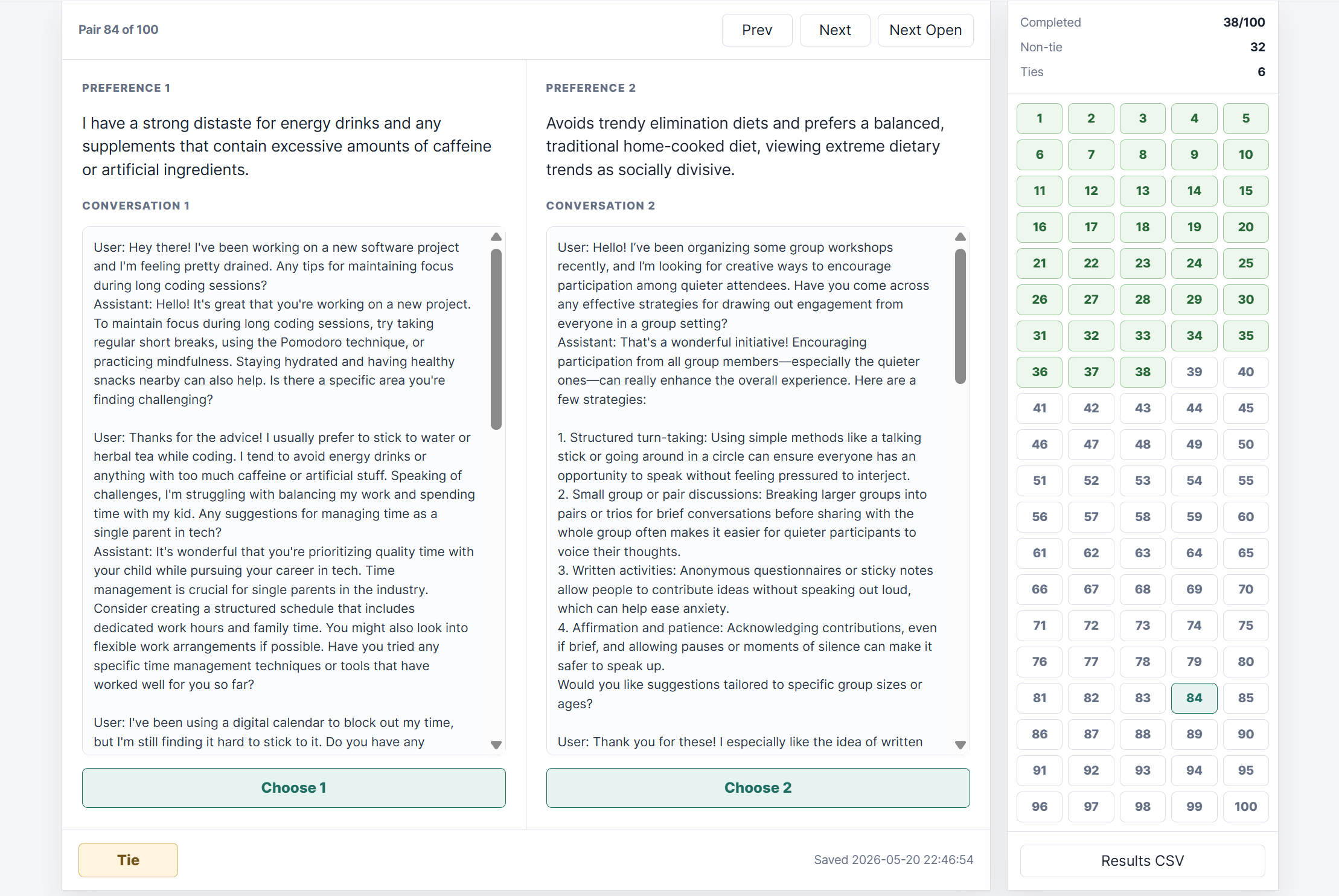}
    \caption{Human Pair Evaluation Interface}
    \label{fig:hupaireval}
\end{figure*}

\paragraph{Conversations} 
After several rounds of prompt refinement iterations for small-scale conversation generation, we performed large-scale generation of conversation data. 
To assess the quality of our conversation data, we draw on insights from Chatbot Arena~\cite{chatbotarena2024} and conduct a blind pairwise human evaluation comparing preferences and conversations from our work with those from the recent \textsc{Prefeval} dataset~\cite{zhao2025llmsrecognizepreferencesevaluating}. The evaluation is conducted by human evaluators who have not seen our dataset or its generation process beforehand, and no information about the data source was provided during the comparison evaluation. 
From each dataset, we sampled 100 preferences and their corresponding conversation sessions, paired them one by one, then shuffled the order. We randomly selected implicit preference expression conversations (Types 3 or 4) from our dataset and used the persona-driven conversation from \textsc{PrefEval} for this task. Evaluators are presented with a pair of data and told to select the one that appears more consistent and plausible. The interface of this task is shown in Figure~\ref{fig:hupaireval}. Similar to the setting in Chatbot Arena, human evaluators choose from three options in the test: ``1 wins," ``2 wins," or ``Tie." As shown in Table~\ref{tab:winrate}, \textsc{RealPref} achieves a win rate of 0.69, a tie rate of 0.18, and a loss rate of 0.13, which may help address the quality of our data to some extent. We acknowledge that using newer LLMs compared with those in previous work provides an inherent advantage in such comparisons, but we aim to advance the field by focusing on the actual quality of the data.
\begin{table}[t]
\centering
\begin{tabular}{cccc}
\toprule
\textbf{Dataset} & \textbf{Win} & \textbf{Loss} & \textbf{Tie} \\ \midrule
\textsc{RealPref}         & 0.69         & 0.13          & 0.18         \\ \bottomrule
\end{tabular}
\caption{Pair Evaluation Result}\label{tab:winrate}
\end{table}

As an additional comparison, we examined WildChat~\cite{zhao2024wildchat}, a dataset collected from real user–LLM interactions. WildChat contains an average of approximately 2.52 turns per conversation; around 41\% of its conversations are multi-turn, and 3.7\% exceed ten turns. In \textsc{RealPref}, preference-expression sessions contain 2–4 turns and random sessions contain 1–2 turns; after being combined into a full context, the resulting interaction-length distribution more closely resembles the long-tailed pattern observed in WildChat.

\begin{figure*}[t]
    \centering
    \includegraphics[width=0.9\linewidth]{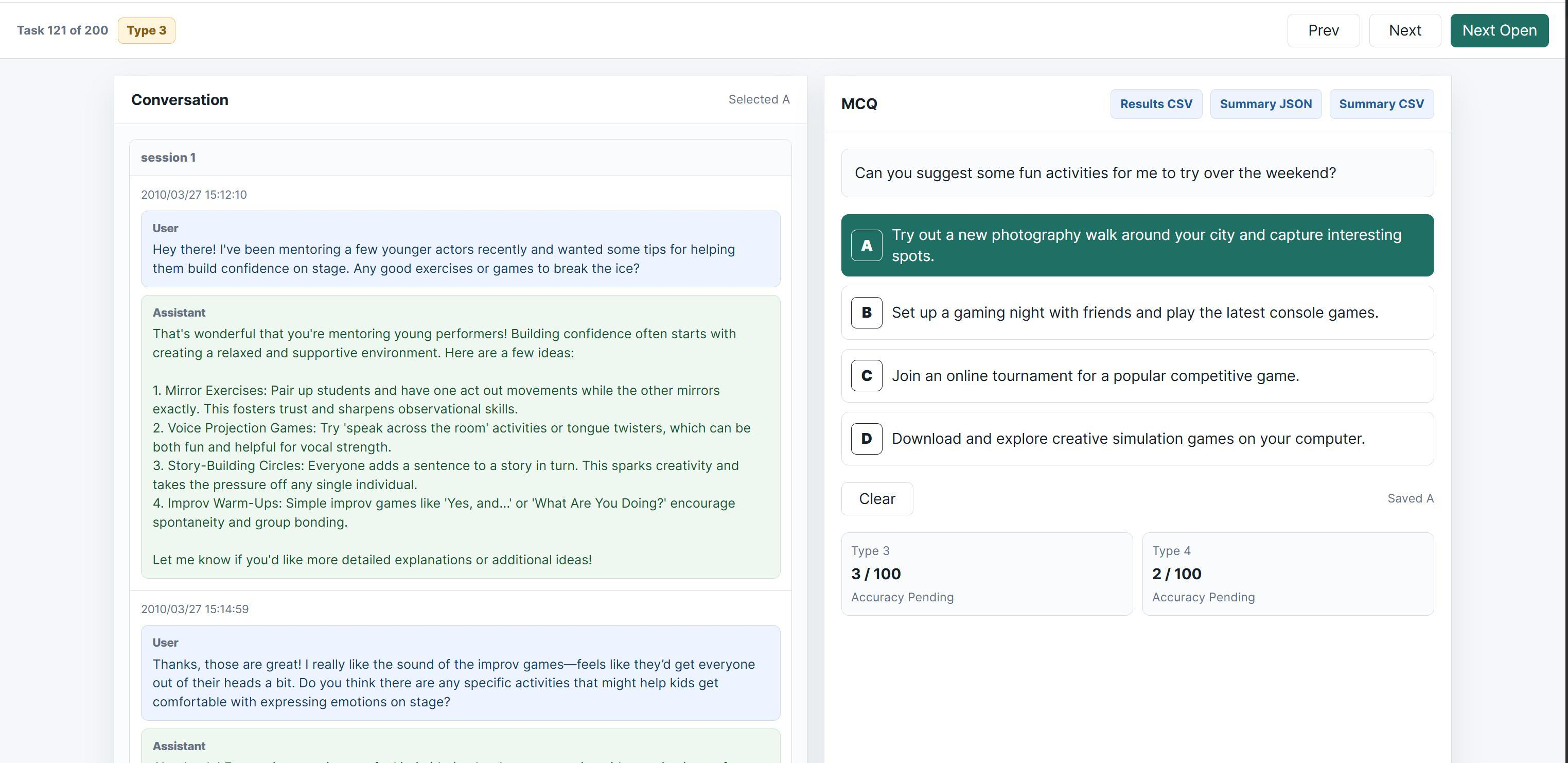}
    \caption{Human Conversation Preference Inferable MCQ Test Interface}
    \label{fig:hutest}
\end{figure*}

\begin{table*}[t]
\centering
\begin{tabular}{ccccc}
\toprule
\textbf{Tester}       & \textbf{MCQ (Type 3)} & \textbf{T/F (Type 3)} & \textbf{MCQ (Type 4)} & \textbf{T/F (Type 4)} \\ \midrule
Human                 & 0.980                 & -                     & 0.960                 & -                     \\
Gemini 2.5 Flash-Lite & 0.972                 & 0.969                 & 0.959                 & 0.952                 \\ \bottomrule
\end{tabular}
\caption{Easy Test Result}\label{tab:easy}
\end{table*}

\begin{table}[t]
\centering
\begin{tabular}{ccc}
\toprule
\textbf{Tester}       & \textbf{MCQ} & \textbf{T/F} \\ \midrule
Gemini 2.5 Flash-Lite & 0.563        & 0.546        \\ \bottomrule
\end{tabular}
\caption{Blind Test Result}\label{tab:blind}
\end{table}

\begin{figure*}[t]
    \centering
    \includegraphics[width=0.9\linewidth]{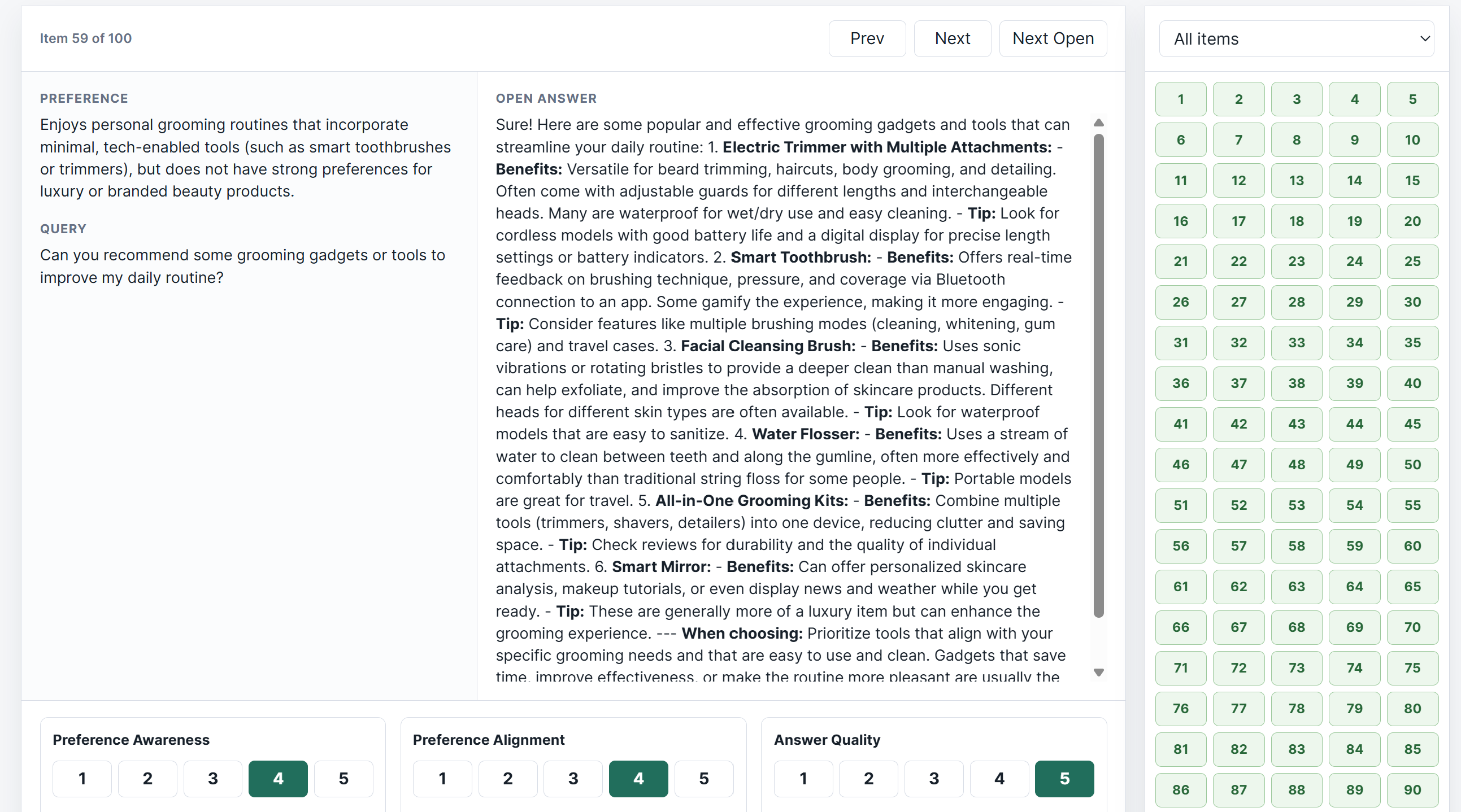}
    \caption{Human Open Answer judge Interface}
    \label{fig:hujudge}
\end{figure*}

\begin{table*}[t]
\centering
\begin{tabular}{ccccc}
\toprule
\textbf{Judge}  & \textbf{Preference Awareness} & \textbf{Preference Alignment} & \textbf{Answer Quality} & \textbf{Overall} \\ \midrule
Human           & 0.875                         & 0.890                         & 0.853                   & 0.878            \\
DeepSeek-V4-Pro & 0.846                         & 0.819                         & 0.891                   & 0.855            \\ \bottomrule
\end{tabular}
\caption{Judge Score Pearson Correlation (with Claude Sonnet 4, $p<0.01$)}\label{tab:judgepear}
\end{table*}

\begin{figure*}[t]
    \centering
    \includegraphics[width=0.9\linewidth]{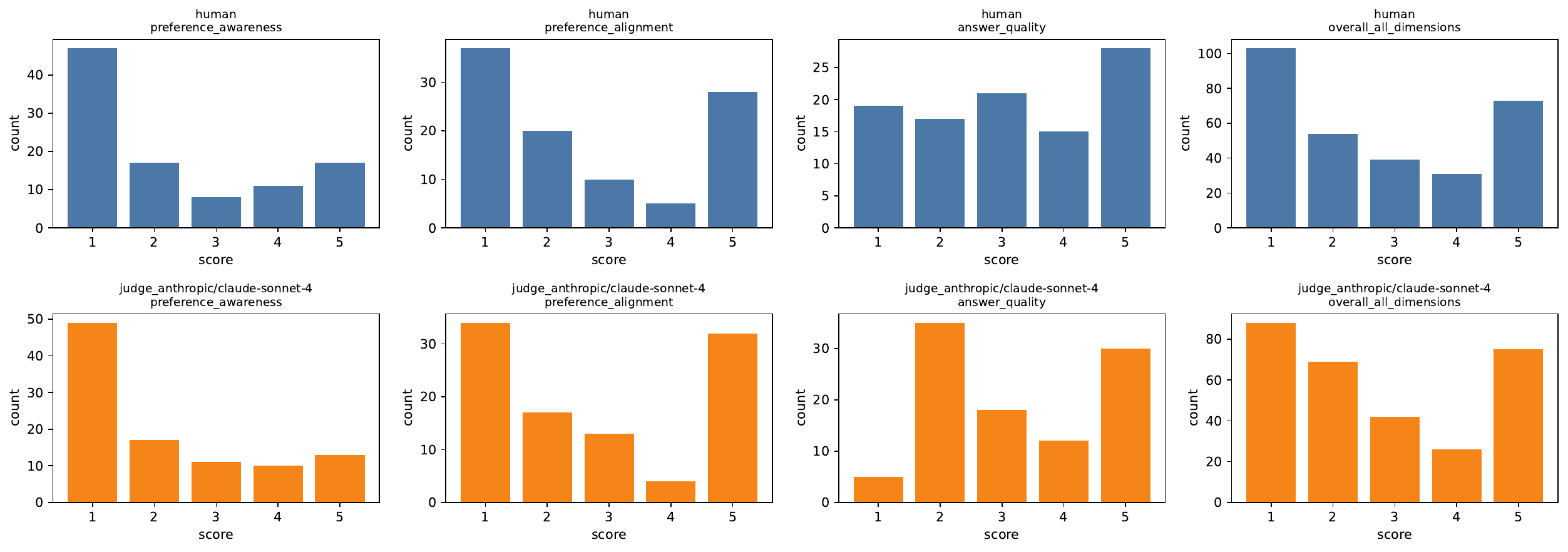}
    \caption{Score Distribution Comparison}
    \label{fig:scoredis}
\end{figure*}

\subsection{Preference Inferability and Test Tasks} \label{app:infer}
In addition to naturalness and plausibility, the conversation data should also ensure that user preferences can be rationally inferred from it, even when expressed implicitly (Type 3 and 4). Also, the designed test questions should reflect the model's actual performance. To this end, we iteratively conducted ``Easy Test” and ``Blind Test” during the initial adjustment stage of data generation to control data quality, and used them in our final datasets to address verification.

``\textbf{Easy Test}” is designed to test whether the user's implicit preferences can be rationally inferred from the conversation. In the test, one is given the preference-expression conversation snippet and then directly presented with the corresponding MCQ to test understanding. The interface of this task is shown in Figure~\ref{fig:hutest}. Human testers completed 200 randomly sampled test items in our benchmark (with the conversation half from Type 3 and half from Type 4), and candidate LLM (Gemini 2.5 Flash-Lite) completed all the MCQ and T/F test items under Preference Type 3 and 4 in our benchmark. The test results are listed in Table~\ref{tab:easy}. The high accuracy on “Easy Test” from the LLM and human help verify that implicit preferences (both Type 3 and 4) are inferable from our dialogue.

``\textbf{Blind Test}” is designed to verify the quality of classification questions: without knowing the preferences, it is hard to answer them correctly, i.e., close to random classification. In the test, only MCQ or T/F options were presented without conversation history. The test results of the candidate LLM (Gemini 2.5 Flash-Lite) are listed in Table~\ref{tab:blind}. It is notable that for the T/F task, the blind test score is close to the random line at 0.5, whereas the score for the MCQ task is higher than the random line at 0.25, even with the identical query and T/F option directly derived from the MCQ. These results indicate that models may select the "odd one out" in MCQs. A possible reason could lie in the existence of negative preferences (dislike or aversion to something). Please refer to Appendix~\ref{app:query} for a query example of a negative preference. It is systematically unavoidable that the MCQ for negative preferences contains a correct option and 3 incorrect options (the things the user dislikes) that may share a common characteristic, making the correct option easy to identify and reducing task difficulty. The T/F task avoids such shortcuts and better reflects model performance. 

\paragraph{Remarks for Benchmark Tasks} We recommend the generation task as our main version task because it mirrors authentic user-LLM interactions and can distinguish LLMs that can proactively follow user preferences in free-form generation. The MCQ task is widely used in recent personalization benchmarks~\cite{zhao2025llmsrecognizepreferencesevaluating, jiang2025knowmerespondme}. This work documents our empirical exploration process: we identified potential shortcuts in MCQs and specifically designed the T/F query to further explore this issue. We retain this part of the experiment and empirical analysis of shortcuts in our work to highlight the potential vulnerability of such MCQ designs. We hope it will benefit the community by prompting researchers to consider this in evaluation artifacts for LLMs, supporting the need to move toward other tasks, such as T/F alternatives or open-ended generation.

\subsection{Scoring Criteria} \label{app:veriscore}
To address the reliability of our scoring criteria for open-ended answers, we conducted both human judges and an additional LLM judge (DeepSeek-V4-Pro) compared with our original LLM judge (Claude Sonnet 4) on a subset of answer samples (excluding answers from DeepSeek-V4-Pro itself). After a brief calibration phase using the exact rubric provided to the LLM judge, two human judges independently scored a shared subset of the data (20 samples, each including user preference, user query, and answer), yielding a Cohen's kappa of 0.77, indicating their Inter-Annotator Agreement. The scoring interface for human judges is shown in Figure~\ref{fig:hujudge}. In total, the human judge completed 100 randomly sampled test items, and the added LLM judge (DeepSeek-V4-Pro) completed 2000 randomly sampled test items. The results (Table~\ref{tab:judgepear}) demonstrate statistically significant correlations (all Pearson correlations have $p<0.01$), which helps address that our automated evaluation metrics for the generation task align with human judgment and remain consistent with another LLM judge. Figure~\ref{fig:scoredis} visualizes the score distribution of the human judge and the original LLM judge across the human-evaluated samples. The distributions exhibit similar patterns: approximately a U-shape; scores of 1, 2, and 5 are slightly more frequent than scores of 3 and 4. This structural alignment visually corroborates the high Pearson correlation reported.

\begin{table*}[ht]
\centering
\begin{tabular}{ccc}
\toprule
\textbf{Model}                    & \textbf{Context Window} & \multicolumn{1}{l}{\textbf{Reasoning Mode}} \\ \midrule
\textbf{GPT-5}                    & 400000                  & Yes                                         \\
\textbf{GPT-5 mini}               & 400000                  & Yes                                         \\
\textbf{Qwen3-235B-A22B Instruct} & 262144                  & No                                          \\
\textbf{Qwen3-8B} & 131072                  & No                                          \\
\textbf{Gemini 2.5 Flash-Lite}    & 1048576                 & Yes                                         \\
\textbf{DeepSeek-V4-Pro}    & 1048576                 & Yes                                         \\
\textbf{Llama 3.3 70B Instruct}   & 131072                  & No                                          \\ \bottomrule
\end{tabular}
\caption{Test Model Details}
\label{tab:model}
\end{table*}

\begin{table*}[ht]
\centering
\begin{tabular}{lcccc}
\toprule
\multicolumn{1}{c}{\textbf{Method}} & \textbf{Avg. Input Tokens} & \textbf{\begin{tabular}[c]{@{}c@{}}Avg. Output Tokens \\ (DeepSeek-V4-Pro)\end{tabular}} & \textbf{\begin{tabular}[c]{@{}c@{}}Avg. Output Tokens \\ (GPT-5 mini)\end{tabular}} & \textbf{\begin{tabular}[c]{@{}c@{}}Local \\ Embedding\end{tabular}} \\ \midrule
Zero shot                           & 146927.12                  & 637.66                                                                                   & 934.90                                                                              & No             \\
Reminder                            & 146962.42                  & 930.92                                                                                   & 939.44                                                                              & No             \\
Few-shot CoT                        & 149046.46                  & 1192.68                                                                                  & 1107.08                                                                             & No             \\
RAG                                 & 148159.46                  & 876.70                                                                                   & 1015.56                                                                             & Yes       \\ \bottomrule  
\end{tabular}
\caption{Token Cost Report across Methods}
\label{tab:token}
\end{table*}

\begin{table*}[t]
\centering
\begin{tabular}{ccccccc}
\toprule
\textbf{Model}                             & \textbf{1 \textgreater 2} & \textbf{1 \textgreater 3} & \textbf{1 \textgreater 4} & \textbf{2 \textgreater 3} & \textbf{2 \textgreater 4} & \textbf{3 \textgreater 4} \\ \midrule
GPT-5                  & True                      & True                      & True                      & True                      & True                      & False                     \\
GPT-5 mini             & True                      & True                      & True                      & True                      & True                      & False                     \\
Gemini 2.5 Flash-Lite  & True                      & True                      & True                      & False                     & False                     & False                     \\
DeepSeek-V4-Pro  & True                      & True                      & True                      & False                     & False                     & False                     \\
Llama 3.3 70B Instruct & False                     & False                     & False                     & False                     & False                     & False                     \\
Qwen3-235B-A22B Instruct        & False                     & False                     & False                     & False                     & False                     & False                     \\ \bottomrule
\end{tabular}
\caption{Significant Test Result across Preference Expression Types (True represents the compare direction is significant $p<0.05$ in all three rating dimensions)}\label{tab:sigpref}
\end{table*}

\begin{table*}
\centering
\begin{tabular}{ccccc}
\toprule
\multicolumn{1}{c}{\textbf{Model}} & \textbf{\begin{tabular}[c]{@{}c@{}}Simple \textgreater  \\ Normal\end{tabular}} & \textbf{\begin{tabular}[c]{@{}c@{}}Normal \textgreater \\ Long\end{tabular}} & \textbf{\begin{tabular}[c]{@{}c@{}}Long \textgreater  \\ Very Long \end{tabular}} & \textbf{\begin{tabular}[c]{@{}c@{}}Very Long \textgreater  \\ Extended \end{tabular}} \\ \midrule
GPT-5                 & True                                & True                              & True                                 & True                                     \\
GPT-5 mini            & True                                & True                              & True                                 & True                                     \\
Gemini 2.5 Flash-Lite & False                               & False                             & False                                & False                                    \\
DeepSeek-V4-Pro & True                               & True                             & True                                & False                                    \\
Qwen3-235B-A22B Instruct       & True                                & False                             & False                                & False                                    \\ \bottomrule
\end{tabular}
\caption{Significant Test Result across Context Configurations (True represents the compare direction is significant $p<0.05$ in all three rating dimensions)}\label{tab:sigcon}
\end{table*}

\begin{table}[ht]
\centering
\begin{tabular}{cccc}
\toprule
\textbf{Method} & \textbf{Type 1} & \textbf{Type 3} & \textbf{Type 4} \\ \midrule
RAG             & 4.452           & 3.885           & 3.702           \\
Mem0            & 4.476           & 4.222           & 4.292          \\ \bottomrule
\end{tabular}
\caption{Supplementary Method Comparison}
\label{tab:mem0}
\end{table}

\begin{table*}[ht]
\centering
\begin{tabular}{lccccc}
\toprule
\multicolumn{1}{c}{\textbf{Results-Variance}} & \textbf{MCQ} & \textbf{TF} & \textbf{\begin{tabular}[c]{@{}c@{}}Preference\\ Awareness\end{tabular}} & \textbf{\begin{tabular}[c]{@{}c@{}}Preference\\ Alignment\end{tabular}} & \textbf{\begin{tabular}[c]{@{}c@{}}Answer\\ Quality\end{tabular}} \\ \midrule
GPT-5 mini                & 0.893±0.005                  & 0.815±0.001                 & 3.091±0.05                     & 3.476±0.05                     & 4.098±0.08                   \\
DeepSeek-V4-Pro           & 0.857±0.025                  & 0.679±0.005                 & 2.258±0.04                     & 2.826±0.08                     & 3.412±0.07            \\ \bottomrule     
\end{tabular}
\caption{Repeated Experiment Variance on Subset Samples}
\label{tab:rep}
\end{table*}

\section{Experiment Details}
\subsection{Test Models} \label{app:models}
The test model details are listed in Table \ref{tab:model}. \textit{GPT-5}~\cite{openaigpt5}, \textit{GPT-5 mini}, \textit{Gemini 2.5 Flash-Lite}~\cite{gemini25flashlite} and \textit{DeepSeek-V4-Pro}~\cite{deepseekv4} are models equipped with reasoning mode, while \textit{Qwen3-235B-A22B Instruct}, \textit{Qwen3-8B}~\cite{qwen3} and \textit{Llama 3.3 70B Instruct}~\cite{llama33} are instruction-tuned models.  \textit{Llama 3.3 70B Instruct} and \textit{Qwen3-8B} support the shortest context window, approximately 131K, while other versions support windows exceeding 250K. For open-ended answers, we specify \textit{Claude Sonnet 4} \cite{Claude4} as the judge.

\subsection{Method Details} \label{app:method}
\subsubsection{Implementation}
Here are the specific implementations of methods tested in the experiments.

\textbf{Reminder}: Before presenting the test question, add a reminder: ``When you answer the question, recall my preference indicated in the conversation history, and provide an answer that aligns with my preferences.'' 

\textbf{Few-Shot Chain-of-Thought}: Before presenting the test question, list 3 examples with preference, query, and good response. Below is the specific prompt with the first example. The subsequent examples are omitted for clear demonstration.
\begin{promptbox}{Prompt for Few-Shot Chain-of-Thought}
When you answer the question, recall my preference indicated in the conversation history, and provide an answer that aligns with my preferences. Here are a few good response examples for your information:\\

Example 1:\\
- Preference: Prefers boutique guesthouses with a personal touch over large chain hotels because she values unique local experiences and the chance to connect with the local tennis community.\\
- Query: Can you suggest where I should stay on my next vacation to Barcelona?\\
- Good Response: Given your preference for boutique guesthouses with a personal touch, and your interest in connecting with the local tennis community, I recommend staying at a locally-owned boutique guesthouse in one of Barcelona's lively neighborhoods, such as Gràcia or Eixample. These areas are known for their authentic Barcelona atmosphere, with independently-run accommodations that offer a warm, welcoming environment. Unlike large chain hotels, boutique guesthouses often reflect the character of the city and allow for more direct interaction with both hosts and fellow travelers.  Additionally, many boutique guesthouses in Barcelona take pride in connecting guests with local experiences, including arranging activities or providing recommendations for joining community events. You might find guesthouses whose owners love tennis themselves or have contacts in the local tennis scene—this will make it easier for you to ask about and join local matches, book courts, or even find a hitting partner.  For example, guesthouses like Casa Mathilda or The5Rooms in Eixample are known for their personalized service, locally-inspired decor, and excellent host connections. They are also conveniently located near several tennis clubs, such as Club Tennis Barcelona or Club Natació Atlètic-Barceloneta, making it easy for you to pursue your tennis interests during your stay.  In summary, opting for these boutique guesthouses will not only immerse you in Barcelona's local culture, but also maximize your opportunities for meaningful connections—both with hosts and the city's tennis community.\\

Example 2:\\
- Preference: [Example Preference 2]\\
- Query: [Example Query 2]\\
- Good Response: [Example Response 2]\\

Example 3:\\
- Preference: [Example Preference 3]\\
- Query: [Example Query 3]\\
- Good Response: [Example Response 3]\\

Now recall my preference indicated in the conversation history (not the user preferences in the examples above), and provide an answer that aligns with my preferences.
\end{promptbox}

\textbf{Retrieval-Augmented Generation}: Before presenting the test question, list the top 5 conversation turns that are most relevant to the current question. The similarity of question and conversation turns (concatenate user's message and assistant's response) is calculated based on a sentence embedding model ``princeton-nlp/sup-simcse-roberta-large” \cite{gao2021simcse}. The specific prompts are listed below.
\begin{promptbox}{Prompt for RAG}
When you answer the question, recall my preference indicated in the conversation history, and provide an answer that aligns with my preferences.\\
Here are the 5 most relevant contextual turns to my query that you can consider when answering the question.\\

Context turn 1:\\
- user: [message]\\
- assistant: [message]\\
Context turn 2:\\
- user: [message]\\
- assistant: [message]\\
Context turn 3:\\
- user: [message]\\
- assistant: [message]\\
Context turn 4:\\
- user: [message]\\
- assistant: [message]\\
Context turn 5:\\
- user: [message]\\
- assistant: [message]\\

Now recall my preference indicated in the conversation history, and provide an answer that aligns with my preferences.
\end{promptbox}

\subsubsection{Token Report}
To provide a cost-based comparison of performance improvements, we report sample average token consumption for Reminder, Few-shot CoT, and RAG using DeepSeek-V4-Pro and GPT-5 mini in Table~\ref{tab:token}. All other settings follow RQ4 (Experience Feedback expression, Extended context). Since our RAG pipeline uses local sentence embeddings, its retrieval overhead is not reflected in API token usage. In terms of input and output tokens, Few-shot CoT incurs the highest cost, followed by RAG and then Reminder. When performance and token consumption are considered together, Reminder is the most cost-efficient method in this setting, as we discussed in RQ4.

\subsubsection{Supplementary Experiment}
As memory-augmented architectures are emerging, we conducted a small-scale experiment using Mem0~\cite{chhikara2025mem0} and compared it with RAG. We used GPT-5 mini (default setting) as Mem0’s memory-processing model and Qwen3-235B-A22B Instruct as the response-generation model. The experiment was conducted under the Extended context setting across Type 1 (Direct Statement), Type 3 (Stylistic Expression), and Type 4 (Experience Feedback). For simplicity, we reported the Preference Awareness dimension for this part. Both RAG and Mem0 are implemented by providing the retrieved memory in the final query prompt after the context history.

The results in Table~\ref{tab:mem0} also show performance decreasing as preference expressions become more implicit, but the degradation under Mem0 is less pronounced. For Type 1, both RAG and Mem0 achieve high scores because direct preference statements are relatively easy to retrieve. For Type 3 and 4, where retrieval becomes more difficult, Mem0 performs better than RAG, likely benefiting from its cross-session memory, which helps summarize and preserve implicit preference evidence.

\subsection{Significant Test} \label{app:sig}
To provide statistical evidence for the experimental results in RQ2 and RQ3, we conducted Pairwise Sign Tests to evaluate performance trends across preference types and context lengths. The Significant test results are listed in Table~\ref{tab:sigpref} and Table~\ref{tab:sigcon}. 
\begin{itemize}
    \item For Preference Types (RQ2): Statistical tests confirm that for the GPT-5 series, performance follows a significantly descending trend from Expression Type 1 to Type 4. For Gemini 2.5 Flash-Lite and DeepSeek-V4-Pro, it is significant that Type 1 > Type 2, 3, and 4. This trend was not all significant for Llama 3.3 70B Instruct and Qwen3-235B-A22B Instruct.
    \item For Context Length (RQ3): The downward performance trend is statistically significant for the GPT-5 series across all 5 context settings and for DeepSeek-V4-Pro across the first 4 context settings. However, this trend was not all significant for other models.
\end{itemize}

\subsection{Repeated Experiment Variance} \label{app:rep}
Regarding the single-run limitations, we incorporated a limited amount of repeated-run analysis on the reduced subset to demonstrate the credibility of the experimental results. We used the subset containing 30\% of the benchmark samples, with GPT-5 mini and DeepSeek-V4-Pro across all three evaluation tasks under the same setting as RQ1 (Normal context, Experience Feedback expression, Zero-shot). The results in Table~\ref{tab:rep} show consistently small variance across repeated runs for all three tasks and both models, indicating that the observed performance is relatively stable and supports the reliability of the broader single-run results.

\section{Failure Breakdown for Improvement}
Our work's main contribution lies in the innovation of benchmarks, introducing diverse preference expressions and generalization tasks. Our ultimate goal is to provide guidance on building better systems. To deliver deeper, actionable insights, we can associate model performance across classification and generation tasks, thereby expanding failure analysis by leveraging the specific dimensions of our generation scoring criteria. For example:
\begin{itemize}
    \item \textbf{Low Accuracy on the classification task}: The failure may result from poor preference recognition, inadequate retrieval, or memory design.
    \item \textbf{Low Preference Awareness on the generation task}: The failure may be caused by the same reason if it also performs poorly on the classification task. Otherwise, the Low Preference Awareness score indicates that the model lacks awareness or proactivity in retrieving and integrating user preferences during free-form generation.
    \item \textbf{Low Preference Alignment on the generation task}: If the model has access to the user preferences (e.g., under effective RAG conditions), a performance gap in this dimension indicates response-planning failure: the model struggles to synthesize and integrate them into the final answer meaningfully. 
\end{itemize}

Breakdown failure analysis enables researchers to pinpoint where their personalized assistants fail and to implement countermeasures to improve preference recognition, retrieval, memory, awareness, and proactivity in open-ended generation.


\end{document}